\documentclass{article} 
\usepackage[preprint]{colm2026_conference}
\usepackage[table]{xcolor}
\usepackage{colortbl}
\pdfoutput=1
\usepackage[most]{tcolorbox}
\usepackage{listings}

\usepackage[most]{tcolorbox}

\usepackage{fontawesome5} 
\usepackage{xurl} 

\usepackage[most]{tcolorbox}
\usepackage{fontawesome5} 

\usepackage{algorithm}      
\usepackage{algpseudocode}  
\usepackage{caption}        

\usepackage{float}    
\usepackage{placeins} 

\usepackage{microtype}
\usepackage{url}
\usepackage[T1]{fontenc}

\usepackage{wrapfig}
\usepackage{graphicx}
\usepackage{subcaption}
\usepackage{algorithm}
\usepackage{amsmath,amssymb}
\usepackage{algorithm}
\usepackage{algpseudocode}

\usepackage[skip=2pt]{caption} 

\usepackage{amssymb,amsmath}
\usepackage{graphicx}

\usepackage[most]{tcolorbox}
\usepackage{xurl} 
\usepackage{listings}

\usepackage{subcaption}
\usepackage{microtype}
\usepackage{xspace}

\usepackage{graphicx}

\usepackage[table]{xcolor}   
\usepackage[most]{tcolorbox} 
\usepackage{tabularx,booktabs,ragged2e}
\usepackage{natbib}
\usepackage{booktabs,siunitx}

\usepackage{setspace} 
\usepackage{booktabs}
\usepackage{tabularx}
\usepackage{ragged2e}
\usepackage{enumitem}
\setlist{resume=never}
\setlist[itemize]{noitemsep, topsep=2pt, leftmargin=*}
\usepackage{multicol}
\usepackage{graphicx}
\usepackage{caption} 

\usepackage{wrapfig}
\usepackage{graphicx}

\usepackage{booktabs,tabularx,siunitx}

\usepackage{booktabs}

\usepackage[bookmarks=false]{hyperref}

\hypersetup{
  colorlinks=true,
  linkcolor=blue,
  citecolor=blue,
  urlcolor=blue
}

\usepackage{lineno}

\definecolor{darkblue}{rgb}{0, 0, 0.5}
\hypersetup{colorlinks=true, citecolor=darkblue, linkcolor=darkblue, urlcolor=darkblue}

\title{AWARE-US: Preference-Aware Infeasibility Resolution in Tool-Calling Agents}

\author{Mehmet Kurmaz\\
Department of Computer Science\\
University College London\\
\texttt{ucabm84@ucl.ac.uk}
}

\begin{document}
\pdfoutput=1
\ifcolmsubmission
\linenumbers
\fi

\maketitle

\begin{abstract}
Tool-calling conversational agents querying structured databases often face two linked failures: underspecification (missing constraints needed for a precise query) and infeasibility (a fully specified query returns an empty set). Prior systems often respond with “no results” or apply ad hoc relaxations, which can violate user intent by discarding highly valued requirements. We cast infeasibility handling as \emph{preference-aware query repair}: when a query is unsatisfiable, the agent should relax the least important constraints. We propose three LLM-based methods to infer relative constraint importance from dialogue: (1) local weighting, (2) global one-shot weighting, and (3) pairwise ranking. Across extensive experiments in car recommendation, the local-weighting method trained with supervised fine-tuning and direct preference optimization best aligns with user preferences (48\%), while global weighting achieves the highest correct-relaxation accuracy (56\%); all three outperform prior infeasibility-resolution basel.

We also introduce AWARE-US, a benchmark of 120+ persona-grounded queries requiring agents to (i) disambiguate a base request via conversation and (ii) resolve infeasibility in a way consistent with persona-implied preferences. For code refer to Github: \url{https://github.com/mhtkrmz/Infeasible-task} and the dataset is available on Hugging Face.\footnote{\url{https://huggingface.co/datasets/MehmetKurmaz/AWARE-US}}
\end{abstract}

\section{Introduction}

\begin{wrapfigure}{r}{0.50\columnwidth}
  \vspace{-10pt} 
  \centering

  \begin{subfigure}[t]{0.485\linewidth}
    \centering
    \includegraphics[width=\linewidth]{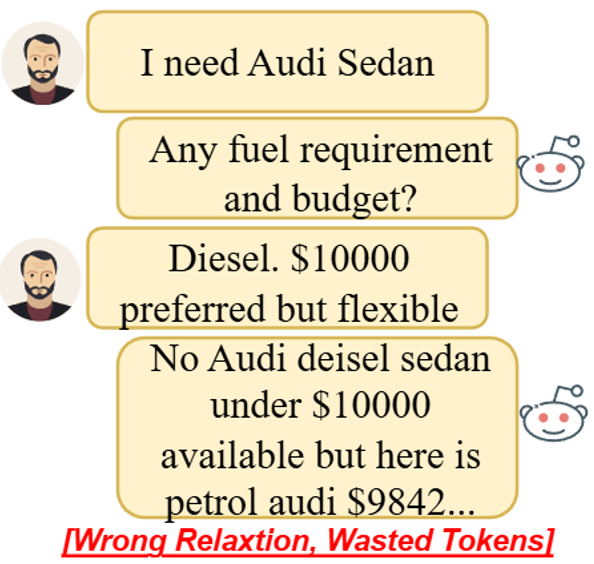}
    \vspace{-8pt} 
    \caption{Undesired Conversation}
    \label{fig:undesired}
  \end{subfigure}
  \hfill
  \begin{subfigure}[t]{0.485\linewidth}
    \centering
    \includegraphics[width=\linewidth]{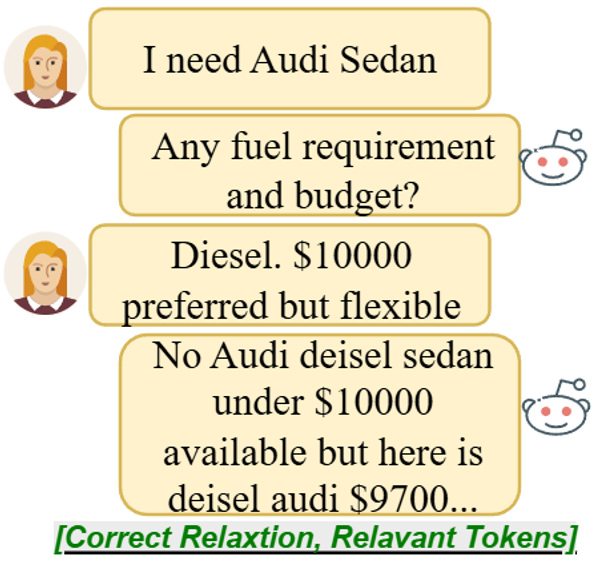}
    \vspace{-8pt}
    \caption{Desired Conversation}
    \label{fig:desired}
  \end{subfigure}

  \vspace{0pt} 
  \caption{Problem Illustration: a) incorrect relaxation wastes tokens and gives irrelevant results; b) correct relaxation yields relevant results and better satisfaction.}
  \label{fig:problem_illustration}
  \vspace{-12pt} 
\end{wrapfigure}

Conversational agents are interactive systems designed to communicate with users in natural language to accomplish goals, ranging from open-ended social chat to goal-directed assistance \citep{wang2025surveyevolutionlanguagemodelbased, acikgoz2024convagents}. A common high-level distinction is between open-domain dialogue and task-oriented dialogue (ToD), where the system helps the user complete a concrete task such as finding an item, booking a service, or answering questions grounded in an underlying knowledge source \citep{oruche2025survey, chen2017survey, wang2025surveyevolutionlanguagemodelbased}.

In task-oriented settings \citep{wang2021task, hosseini2020simple}, the agent’s job is not only to produce fluent responses but to ground language in actions: it must interpret the user’s intent, track constraints over turns (dialogue state tracking), decide what to do next (a dialogue policy), and finally execute a backend action such as a database lookup or API call before generating the response \citep{wang2025surveyevolutionlanguagemodelbased, qin2023endtoendtaskorienteddialoguesurvey}. This view has been central to the ToD literature and motivates why backend connectivity is essential for real utility \citep{qin2023endtoendtaskorienteddialoguesurvey}.

\setlength{\textfloatsep}{1pt}
\setlength{\intextsep}{1pt}
\begin{figure*}[t]
  \centering
  \includegraphics[width=\textwidth]{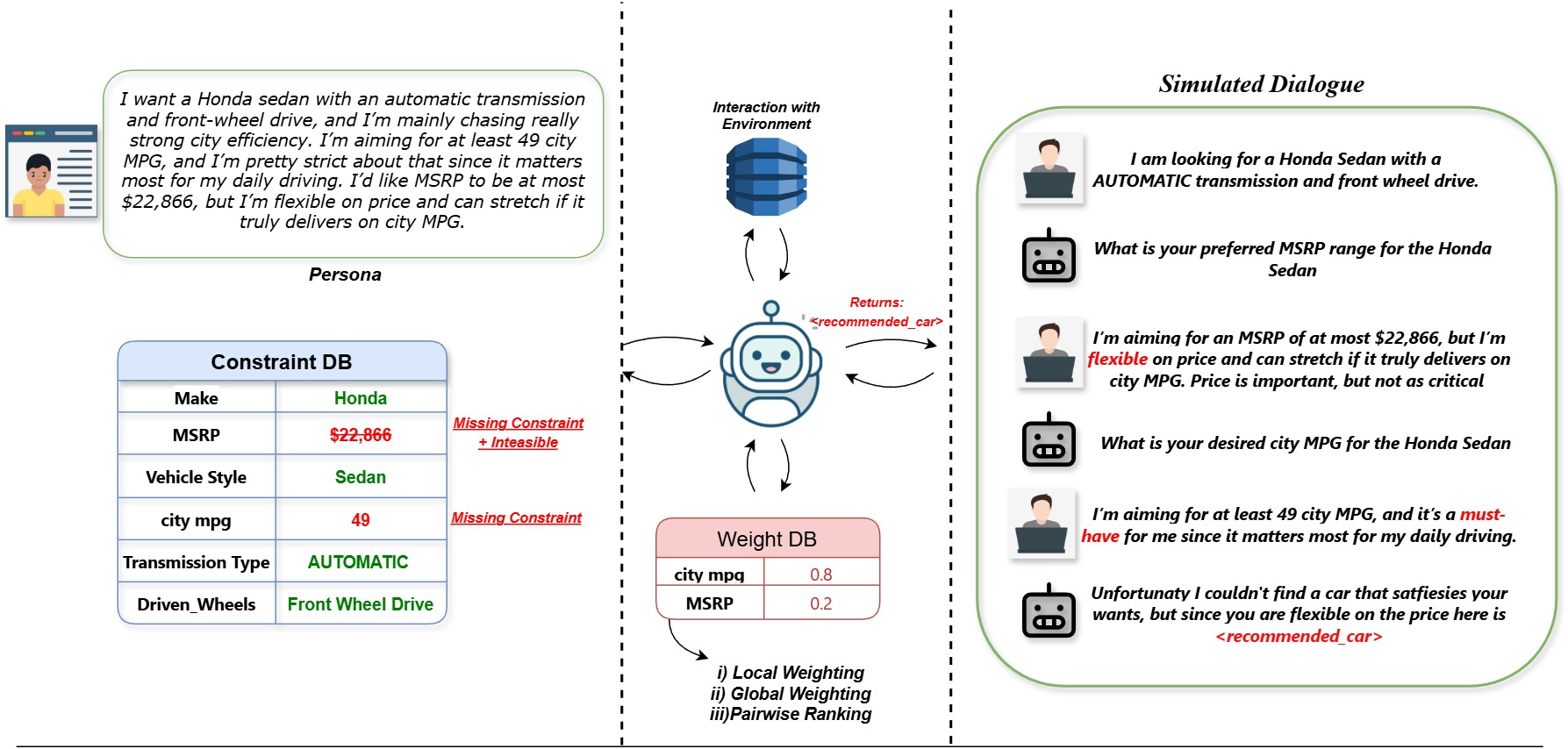} 
  \caption{Overview of our Preference-Aware Resolution Framework: Given a user query, specifying desired attributes and soft preferences, our framework elicits missing constraints via dialogue, extracts the resulting constraint set, and estimates preference weights (local weighting, global weighting, or pairwise ranking). The weighted constraints are used to query the constraint database; if the query is infeasible, the agent relaxes the lowest-importance constraint(s) to recover a feasible candidate and recommends optimized result based on user preferences.}
   \vspace{0.2cm}
  \label{fig:wide}
\end{figure*}

Tool-calling agents that repeatedly query structured backends face two main failure modes: underspecification, when user requests omit constraints needed to form an executable query, and infeasibility, when a fully specified query returns an empty set \citep{zhang2024askbeforeplanproactivelanguageagents, hao2025largelanguagemodelssolve, anonymous2025argus, shim2025tooldialmultiturndialoguegeneration}. These issues are studied heavily in travel planning; for instance, TravelPlanner offers a tool sandbox over ~4M crawled records accessed via six tools and includes 1,225 curated query–plan pairs \citep{xie2024travelplannerbenchmarkrealworldplanning}. Building on this line, Ask-Before-Plan \citep{zhang2024askbeforeplanproactivelanguageagents} addresses infeasibility with a static dependency graph of “indefinite details,” asking clarifications in a fixed topological order; when the environment flags an unfeasible detail, it requests an alternative, but subsequent clarifications follow the predefined structure rather than inferred, user-specific preferences, effectively baking domain assumptions into repair. In contrast, LLMs for Formal Verification \citep{hao2025largelanguagemodelssolve} treats infeasibility as an SMT-solver signal: it encodes intent as constraints, uses an UNSAT result and unsat core to surface concrete failure reasons, and enters an interactive repair loop where the LLM gathers tool evidence, proposes constraint edits, and repeatedly solicits user agree/disagree feedback until satisfiable, with alignment driven largely by continued user input \citep{hao2025largelanguagemodelssolve}. Meanwhile, Argus casts underspecification and infeasibility as tool-call disambiguation evaluated on ClarifyBench \citep{anonymous2025argus}, adding an explicit recovery step that either repairs the call or triggers targeted clarifications, often increasing user back-and-forth.

In light of these limitations, we formulate a new task: Preference-Aware Resolution. The goal is to handle infeasible queries, also called "the empty-set problem" in database systems \citep{mottin2013probabilistic, mottin2014iqr} in a preference-aware way. When an LM agent encounters an infeasible query, it should relax the constraint that is least important to the user, using preference cues expressed throughout the conversation, rather than relying on fixed domain priorities or repeated trial-and-error clarification. This moves tool-calling agents toward more negotiation-capable behavior, where repair decisions reflect what the user values most.

Preference-Aware Resolution goes beyond choosing which constraint to relax: after repair yields a feasible set, the agent should rank and select results using the inferred preference structure, treating strong preferences as objectives and weaker ones as tie-breakers, so recommendations remain consistent with the user’s priorities within the feasible region. To study this, we introduce AWARE-US, a persona-grounded, multi-turn car-query benchmark that requires agents to clarify underspecified requests and resolve infeasibility in line with implied preferences, and we propose three LLM-based importance inference methods: (i) local weighting, (ii) global one-shot weighting, and (iii) pairwise ranking.

In summary, our contributions are:
\begin{itemize}
  \item We introduce \textit{Preference-Aware Resolution}, a new problem setting that studies how conversational LLM agents should handle infeasible (empty-set) queries by relaxing constraints in a way that preserves user intent.
  \item We propose three LLM-based frameworks for inferring relative constraint importance from dialogue and using it to guide query repair: local weighting, global one-shot weighting, and pairwise ranking.
  \item We construct \textbf{AWARE-US}, the first benchmark designed specifically for persona-grounded underspecification + infeasibility resolution in structured querying, requiring agents to both disambiguate requests via conversation and resolve infeasibility consistent with implied preferences.
  \item We propose a training strategy that combines supervised fine-tuning (SFT) \citep{ouyang2022traininglanguagemodelsfollow} and direct preference optimization (DPO) \citep{rafailov2024directpreferenceoptimizationlanguage} for this problem, and conduct extensive evaluations and analyses across diverse settings to validate the effectiveness of our approaches.
\end{itemize}

\section{Related Work}
\textbf{Language Model Based Agents} Language agents have the capability to decompose complex tasks and arrive at solutions through a series of reasoned actions powered by Large language models (LLMs) \citep{xi2023risepotentiallargelanguage, deng2024large, Wang_2024} to conduct reasoning \citep{yao2022react, shinn2023reflexionlanguageagentsverbal}, tool use \citep{qin2023toolllmfacilitatinglargelanguage} , memory storage and retrival\citep{wen2024diluknowledgedrivenapproachautonomous}.  A representative line is ReAct, which interleaves reasoning traces with environment actions to improve factuality and robustness in interactive settings \citep{yao2023reactsynergizingreasoningacting}. Tool-use capability is also studied through training methods and benchmarks. Toolformer proposes  self-supervised data generation to teach models when/how to call tools \citep{schick2023toolformerlanguagemodelsteach}. Gorilla fine-tunes LMs to produce accurate API calls, often with retrieval to ground the call in documentation \citep{patil2023gorillalargelanguagemodel}.  Evaluation has consolidated around tool-use benchmarks such as API-Bank (tool-augmented dialogues with annotated API calls) and ToolBench (large-scale tool manipulation evaluation) \citep{qin2023toolllmfacilitatinglargelanguage, li2023apibankcomprehensivebenchmarktoolaugmented}. However, most prior work assumes user instructions are clear enough for language agents to carry out, but in real settings instructions are often brief, which can introduce ambiguity and uncertainty.

\textbf{Preference elicitation with agent} Ambiguity is a core challenge for conversational agents because unclear requests often trigger extra clarification turns, increasing interaction cost \citep{andukuri2024stargateteachinglanguagemodels,   zhang2024askbeforeplanproactivelanguageagents, anonymous2025argus}. Prior work on clarification question generation studies how agents should ask targeted questions to resolve uncertainty efficiently \citep{li2023apibankcomprehensivebenchmarktoolaugmented, zhang2023clarifynecessaryresolvingambiguity}. In most agent querying settings, “ambiguous” is typically treated as underspecified \citep{zhang2024askbeforeplanproactivelanguageagents, hao2025largelanguagemodelssolve, andukuri2024stargateteachinglanguagemodels} missing constraints/information needed to form a precise action/query so existing systems emphasize information-seeking behaviors (slot filling and constraint elicitation). This view is also reflected in conversational semantic parsing benchmarks such as SParC and CoSQL \citep{yu2019cosqlconversationaltexttosqlchallenge, yu2019sparccrossdomainsemanticparsing}, which require context-dependent query construction. Preference elicitation is closely related, especially in conversational recommender systems, where dialogue refines user preferences and trade-offs \citep{Jannach_2021}.
By contrast, less work addresses fully specified but infeasible queries, where the agent must choose what to relax in line with user priorities rather than fixed heuristics or trial-and-error though database systems have studied this setting.

\textbf{Infeasibility Resolution in Database Systems} Empty answers in structured search are commonly handled via query relaxation, dropping or weakening conjunctive predicates until satisfiable \citep{mottin2013probabilistic, ikeda2024efficient, mottin2014iqr}. Prior work formulates relaxation as sequential optimization trading off user acceptance and answer utility, but the search over relaxation sequences can be exponential \citep{mottin2014iqr}. IQR uses a relaxation tree with iterative yes/no feedback \citep{mottin2014iqr}, and later methods improve efficiency and result quality (e.g., MMR for diversity) while noting enumerating relaxed queries is infeasible \citep{ikeda2024efficient}. These approaches generally use fixed objectives/acceptance models rather than inferring user-specific constraint importance from dialogue.

\vspace{-0.51cm}


\section{AWARE-US: Dataset Construction}
\vspace{-0.21cm}

AWARE-US benchmarks \textbf{preference-aware query disambiguation and infeasibility resolution} for structured retrieval over a tabular car catalog \citep{cooperunion_car_features_msrp_2016}. Each instance is designed so an agent must (i) elicit missing, persona-backed constraints via dialogue and (ii) repair an infeasible query by relaxing the \emph{least important} constraint (not a fixed heuristic). Generation pseudocode appears in Appendix~\ref{Algos}

\textbf{Queries as base slice + preferences.}
We write a query as $q = B \wedge \bigwedge_{i=1}^{k} a_i$, where the \emph{base slice} $B$ fixes four structural attributes (\texttt{Make}, \texttt{Vehicle Style}, \texttt{Transmission Type}, \texttt{Driven\_Wheels}) and is sampled to yield a moderately sized subset for meaningful trade-offs. The additional constraints $\{a_i\}_{i=1}^k$ encode user-specific preferences, drawn from (a) categorical equalities (e.g., \texttt{Model}, \texttt{Vehicle Size}, \texttt{Engine Fuel Type}) and (b) numeric thresholds (e.g., min year/mpg, max MSRP). To keep instances realistic, categorical values must have support within $B$, and numeric thresholds are sampled from within-slice quantiles plus near-extremes to make infeasibility attainable without being implausible; we also filter ``loose'' cases where a single relaxation returns an overly large candidate set.

\textbf{Controlled infeasibility regimes.}
We enforce empty results for the fully specified query, i.e., $R(B,\{a_i\}) = \emptyset$ where $R(\cdot)$ is the set of catalog rows satisfying the conjunction. For $k{=}4$, we generate two regimes: (S1) \emph{unique drop-one repair}, where $\exists! j$ such that $R(B,\{a_i\}_{i\neq j}) \neq \emptyset$, and (S2) \emph{any-one repair}, where $\forall j,\; R(B,\{a_i\}_{i\neq j}) \neq \emptyset$ (a minimal-unsatisfiable-style setting \citep{reiter1987theory,liffiton2008algorithms,bacchus2015using}). Feasibility checks are fast: each predicate is a precomputed bitset over $B$, so conjunctions reduce to bitwise intersections, enabling guided search over infeasible quartets.

\textbf{Oracle preferences and gold targets.}
We sample nonnegative weights $\{w_i\}$ and normalize ($w_i \leftarrow w_i/\sum_\ell w_\ell$) to define an oracle importance profile; the intended relaxation is $j^\star=\arg\min_i w_i$. After relaxing $a_{j^\star}$, we choose a gold recommendation
$x^\star=\arg\max_{x \in R(B,\{a_i\}_{i\neq j^\star})} S(x)$,
where $S$ rewards satisfying higher-weight constraints and uses normalized utilities for numeric attributes when applicable. This yields two targets per instance: the \textbf{relaxed constraint} and the \textbf{recommended record}, isolating the core question: from dialogue alone, can the agent infer which constraint is most negotiable and relax it?

\textbf{Personas.}Each instance includes a short first-person persona that expresses the motivations and trade-offs implied by the oracle weights. Personas are generated with \textbf{GPT-5} \citep{openai_gpt5_docs} few-shot (N = 3), using prompts that encourage internal faithful constraint coverage \citep{jandaghi2024faithful}. 

\textbf{Benchmark settings.}
AWARE-US comprises three controlled settings: S1 (4 constraints, \emph{unique} drop-one repair), S2 (4 constraints, \emph{any-one} repair), and S3 (2 constraints, any-one repair). Table~\ref{tab:awareus-settings} reports sizes.

\begin{table}[t]
\centering
\small
\setlength{\tabcolsep}{4pt}
\renewcommand{\arraystretch}{1.1}
\begin{tabular}{p{0.55\columnwidth}cc}
\toprule
\textbf{Setting} & \textbf{\# additional} & \textbf{\# instances} \\
& \textbf{constraints} & \\
\midrule
S1: $k{=}4$ (unique drop-one repair)    & 4 & 41 \\
S2: $k{=}4$ (any-one repair)            & 4 & 41 \\
S3: $k{=}2$ (any-one repair)            & 2 & 40 \\
\bottomrule
\end{tabular}
\caption{AWARE-US settings and sizes.}
\label{tab:awareus-settings}
\end{table}

\section{Infeasibility Task}
\vspace{-0.1cm}
\label{infeasibility task}
We study \textbf{preference-aware infeasibility resolution} for the failure modes described above. Let $D$ denote the database, and let a user request induce constraints $C=C_{\text{base}}\cup C_{\text{add}}$, where $C_{\text{base}}$ defines a base slice and $C_{\text{add}}$ are additional preferences elicited through dialogue. The executor returns the feasible set $F(C)={x\in D \mid x \models C}$. If $F(C)\neq\varnothing$, the agent recommends an item from $F(C)$; otherwise ($F(C)=\varnothing$) it must \textbf{repair} the query by relaxing some subset of $C_{\text{add}}$ while preserving the user’s preference ordering \citep{koudas2006relaxing}. Task prompts appear in Appendix~\ref{taskprompts} and an example dialogue in Appendix~\ref{exampledialog}.

\textbf{Baselines: greedy query relaxation and solver-style verification.}
\label{greedy}
We introduce two baselines. \textbf{(1) Greedy query relaxation \citep{corella2009brief}}: After eliciting slot values, it executes the full conjunctive query and, if UNSAT, drops one additional constraint at a time preferring the drop that yields the smallest non-empty candidate to restore feasibility; it then recommends an item closest to the original constraints. \textbf{(2) Solver-style satisfiability + repair \citep{hao2025largelanguagemodelssolve}} reimplements the ``LLM + verifier/solver'': the LLM produces structured constraints and a deterministic executor checks satisfiability and exposes systematic repairs via an UNSAT witness/unsat-core signal. This baseline reliably detects infeasibility, but it does not specify which constraint to relax in a preference-aligned way; it therefore serves as a comparison point for preference-agnostic repair.

\textbf{Preference-aware repair: three variants.}
All methods share the same outer loop: (i) run clarification to instantiate $C_{\text{add}}$; (ii) test feasibility; (iii) if infeasible, relax constraints in a preference-guided order until feasible; and (iv) if multiple feasible items remain, rank candidates by a soft satisfaction score.

We explore three ways to infer the relative importance of constraints $c_i \in C_{\text{add}}$:

\noindent\textbf{Local weighting.}
For each slot $i$, the agent asks one clarification question and receives a slot-specific answer.
From that \emph{local} exchange it extracts (i) a concrete constraint $c_i$ and (ii) an importance score $w_i \in [0,1]$ inferred from the user’s language in that same turn. If the full set is infeasible, the repair step relaxes constraints in increasing importance, i.e., relax the smallest $w_i$ first until the feasible set becomes non-empty.

\noindent\textbf{Global one-shot weighting.}
The agent extracts constraints $C={c_i}{i=1}^m$ during dialogue, but assigns importance only after observing the full transcript $S$. In a single global pass it computes weights ${w_i}{i=1}^m$ with $w_i!\ge!0$ and $\sum_i w_i!=!1$, conditioning jointly on $(S,C)$ to capture cross-turn trade-offs. If the full query is infeasible, it relaxes exactly one constraint, selecting $c_{j}$ where $j=\arg\min_i w_i$.

\noindent\textbf{Pairwise ranking \citep{qin2024largelanguagemodelseffective}.}
Rather than estimating calibrated weights, we treat the LLM as a pairwise comparator. For each pair $(c_i,c_j)\in C^2$ it predicts $c_i \succ c_j$ (or $c_j \succ c_i$) conditioned on the dialogue $S$, using an order-swapped query to reduce positional bias. We then aggregate pairwise outcomes into a score $s_i$ (wins, with ties split) and induce a total order by sorting $s_i$. Repair relaxes the lowest-ranked constraint first, i.e., $c_{\arg\min_i s_i}$.

We also introduce a hybrid, \textbf{local weighted + greedy}: if relaxing the least-important constraint still yields UNSAT, greedy method is applied outlined earlier. We include this to disentangle preference-ranking errors from feasibility/search failures.

\section{Two-Stage Training for Preference-Aware Repair}
\label{training}
To test whether AWARE-US supports learnable preference-aware repair, we adopt a standard SFT$\rightarrow$DPO recipe \citep{ouyang2022traininglanguagemodelsfollow,rafailov2024directpreferenceoptimizationlanguage,dettmers2023qloraefficientfinetuningquantized,tunstall2023zephyr} on Qwen/Qwen3-8B \citep{yang2025qwen3technicalreport}. Each benchmark item is indexed by \((q,u)\), where \(q\) is the base query sentence and \(u\) is the persona; we deduplicate to obtain a set \(\mathcal{I}\subseteq \mathcal{Q}\times\mathcal{U}\). Using a fixed seed, we form an item-level split \(\mathcal{I}=\mathcal{I}_{\mathrm{tr}}\cup\mathcal{I}_{\mathrm{te}}\) with \(|\mathcal{I}_{\mathrm{tr}}|:|\mathcal{I}_{\mathrm{te}}|=80:20\), spanning all settings.

\textbf{Stage 1 (SFT).} We fine-tune a persona-conditioned Roleplayer with QLoRA \citep{dettmers2023qloraefficientfinetuningquantized} to answer slot-clarification questions with a single schema-conforming JSON. Let \(\mathcal{S}\) be the slot set and \(C_i\subseteq \mathcal{S}\times\{=,\le,\ge\}\times\mathcal{V}\) the additional constraints for item \(i\). For each \(s\in\mathcal{S}\) we construct a prompt \(x_{i,s}\) (containing \((q_i,u_i)\), a question about \(s\), and the schema) and a completion \(y_{i,s}\) encoding a decision--value pair: a binary indicator \(b_{i,s}\in\{0,1\}\) and, when \(b_{i,s}=1\), a predicate \((o_{i,s},v_{i,s})\in\{=,\le,\ge\}\times\mathcal{V}\). Positives enforce correctness by setting \(b_{i,s}=1\) and \((o_{i,s},v_{i,s})=(o,v)\) for the unique \((s,o,v)\in C_i\); negatives sample \(s\notin \{s':(s',*,*)\in C_i\}\) and set \(b_{i,s}=0\). If item weights \(w_i(s,o,v)\in[0,1]\) are available, we attach an ordinal importance label via a fixed monotone map \(\phi\) (so \(w\le w'\Rightarrow \phi(w)\le \phi(w')\)), ensuring the schema expresses relative importance without changing the underlying constraint semantics.

Training uses completion-only maximum likelihood (prompt masking), so the model learns the conditional distribution of schema-valid answers given the persona and slot query:
\[
\max_{\theta}\ \sum_{(x,y)\in \mathcal{D}_{\text{SFT}}}\ \sum_{t\in \text{completion}(y)} \log \pi_{\theta}\!\left(y_t \mid x, y_{<t}\right),
\]
where \(\mathcal{D}_{\mathrm{SFT}}=\{(x_{i,s},y_{i,s})\}\). Intuitively, this objective factorizes the supervision into (i) learning the \emph{abstain vs.\ commit} decision \(b_{i,s}\) (preference present or not) and (ii) when committing, emitting the correct predicate \((o_{i,s},v_{i,s})\) and ordinal importance consistent with \(u_i\); masking the prompt tokens concentrates gradient signal on producing stable, schema-conforming completions.

\textbf{Stage 2 (DPO).} SFT teaches the model to \emph{represent} user preferences, but not to \emph{act} on them reliably when the full constraint set is infeasible. We therefore apply Direct Preference Optimization (DPO) \citep{rafailov2024directpreferenceoptimizationlanguage} to learn a one-step repair policy. For each item \(i\), let \(C_i=\{c_{i,1},\dots,c_{i,m}\}\) be the additional constraints and assume the full query is unsatisfiable, \(F(C_i)=\varnothing\) (given the fixed base constraints). A repair corresponds to choosing exactly one constraint \(c\in C_i\) to drop; we encode this choice as a completion \(y\) with \(\operatorname{drop}(y)\in C_i\).

We construct preference triples \((x,y^{+},y^{-})\), where \(x\) contains the query/persona context plus an elicitation dialogue and the fact that \(F(C_i)=\varnothing\); \(y^{+}\) drops the oracle-consistent constraint \(c_i^\star\), and \(y^{-}\) drops an alternative \(c\neq c_i^\star\). Rejected choices are sampled to be plausible (and often difficult) by prioritizing alternatives that still yield infeasibility after dropping \(c\), or otherwise yield relatively small feasible sets, while remaining preference-inconsistent.

DPO trains \(\pi_\theta\) against a frozen reference \(\pi_{\mathrm{ref}}\) (initialized from the SFT model) by increasing the relative log-likelihood gap between preferred and rejected repairs:
\[
\begin{aligned}
\mathcal{L}_{\text{DPO}}(\theta)
&= -\mathbb{E}_{(x,y^{+},y^{-})\sim \mathcal{D}_{\text{DPO}}}
\left[
\log \sigma\!\Big(
\beta\big(
\Delta_{\theta}(x) - \Delta_{\mathrm{ref}}(x)
\big)
\Big)
\right],\\
\Delta_{\theta}(x)
&= \log \pi_{\theta}(y^{+}\!\mid x)\;-\;\log \pi_{\theta}(y^{-}\!\mid x),
\end{aligned}
\]
where \(\beta\) controls preference strength, and \(\Delta_{\mathrm{ref}}\) is defined analogously under \(\pi_{\mathrm{ref}}\). Training pseduocode appears in Appendix \ref{Algos}.

\section{Evaluation and Experimental Results}
We evaluate preference-aware infeasibility handling along three dimensions: (i) dialogue completion, (ii) feasibility/planning outcome, and (iii) oracle agreement with the intended relaxation and recommendation. For each model and method, we run the full end-to-end pipeline over all benchmark instances and report aggregated scores. For oracle-based metrics, we compute agreement only over the subset of instances where the comparison is well-defined. Full metric definitions and computation details are provided in Appendix \ref{Eval}.


\definecolor{HeaderGray}{gray}{0.93}
\definecolor{RowGray}{gray}{0.97}
\newcommand{\best}[1]{\cellcolor{green!15}\textbf{#1}}
\renewcommand{\second}[1]{\cellcolor{yellow!15}#1}

\begin{table*}[t]
\centering
\small
\renewcommand{\arraystretch}{1.18}
\setlength{\tabcolsep}{5pt}
\sisetup{
  table-number-alignment = center,
  round-mode = places,
  round-precision = 1
}
\resizebox{\textwidth}{!}{%
\begin{tabular}{l
S[table-format=1.3]
S[table-format=3.1]
S[table-format=3.1]
S[table-format=3.1]
S[table-format=3.1]
S[table-format=3.1]
S[table-format=3.1]
S[table-format=3.1]
}
\toprule
& \multicolumn{2}{c}{\cellcolor{red!15}\textbf{Dialogue Completion}} &
\multicolumn{4}{c}{\cellcolor{blue!15}\textbf{Planning Outcome}} &
\multicolumn{2}{c}{\cellcolor{green!15}\textbf{Oracle Agreement}} \\
\cmidrule(lr){2-3}\cmidrule(lr){4-7}\cmidrule(lr){8-9}
\textbf{Approach} &
\cellcolor{red!10}\textbf{Avg Cons.} &
\cellcolor{red!10}\textbf{Slot Comp.} &
\cellcolor{blue!10}\textbf{SAT no relax} &
\cellcolor{blue!10}\textbf{SAT after relax} &
\cellcolor{blue!10}\textbf{UNSAT} &
\cellcolor{blue!10}\textbf{Reco rate} &
\cellcolor{green!10}\textbf{Relax match} &
\cellcolor{green!10}\textbf{Car match} \\
\midrule

Greedy (\small{Qwen3-8B}) &
\cellcolor{red!6}\textbf{4.0} & \cellcolor{red!6}89.6 &
\cellcolor{blue!6}36.6 & \cellcolor{blue!6}63.4 & \cellcolor{blue!6}\textbf{0.0} & \cellcolor{blue!6}\textbf{100.0} &
\cellcolor{green!6}43.9 & \cellcolor{green!6}19.5 \\
Formal (\small{Qwen3-8B}) &
\cellcolor{red!6}3.9 & \cellcolor{red!6}\textbf{97.6} &
\cellcolor{blue!6}\textbf{4.9} & \cellcolor{blue!6}\textbf{95.1} & \cellcolor{blue!6}\textbf{0.0} & \cellcolor{blue!6}\textbf{100.0} &
\cellcolor{green!6}14.6 & \cellcolor{green!6}4.9 \\
Global (\textsc{Qwen3-8B}) &
\cellcolor{red!6}3.683 & \cellcolor{red!6}92.1 &
\cellcolor{blue!6}14.6 & \cellcolor{blue!6}80.5 & \cellcolor{blue!6}4.9 & \cellcolor{blue!6}95.1 &
\cellcolor{green!6}\textbf{65.9} & \cellcolor{green!6}24.4 \\
Weighted (\textsc{Qwen3-8B}) &
\cellcolor{red!6}3.854 & \cellcolor{red!6}96.3 &
\cellcolor{blue!6}9.8 & \cellcolor{blue!6}56.1 & \cellcolor{blue!6}34.1 & \cellcolor{blue!6}65.9 &
\cellcolor{green!6}46.3 & \cellcolor{green!6}\textbf{43.9} \\
PRP (\textsc{Qwen3-8B}) &
\cellcolor{red!6}3.561 & \cellcolor{red!6}89.0 &
\cellcolor{blue!6}31.7 & \cellcolor{blue!6}56.1 & \cellcolor{blue!6}12.2 & \cellcolor{blue!6}87.8 &
\cellcolor{green!6}48.8 & \cellcolor{green!6}17.1 \\
\midrule

Greedy (\textsc{Gemma-2-9b-it}) &
\cellcolor{red!6}\textbf{4.0} & \cellcolor{red!6}96.3 &
\cellcolor{blue!6}12.2 & \cellcolor{blue!6}87.8 & \cellcolor{blue!6}\textbf{0.0} & \cellcolor{blue!6}\textbf{100.0} &
\cellcolor{green!6}61.0 & \cellcolor{green!6}24.4 \\
Formal (\textsc{Gemma-2-9b-it}) &
\cellcolor{red!6}3.902 & \cellcolor{red!6}97.6 &
\cellcolor{blue!6}\textbf{0.0} & \cellcolor{blue!6}\textbf{100.0} & \cellcolor{blue!6}\textbf{0.0} & \cellcolor{blue!6}\textbf{100.0} &
\cellcolor{green!6}12.2 & \cellcolor{green!6}4.9 \\
Global (\textsc{Gemma-2-9b-it}) &
\cellcolor{red!6}\textbf{4.0} & \cellcolor{red!6}\textbf{100.0} &
\cellcolor{blue!6}\textbf{0.0} & \cellcolor{blue!6}80.5 & \cellcolor{blue!6}19.5 & \cellcolor{blue!6}80.5 &
\cellcolor{green!6}\textbf{63.4} & \cellcolor{green!6}24.4 \\
Weighted (\textsc{Gemma-2-9b-it}) &
\cellcolor{red!6}3.878 & \cellcolor{red!6}97.0 &
\cellcolor{blue!6}9.8 & \cellcolor{blue!6}56.1 & \cellcolor{blue!6}34.1 & \cellcolor{blue!6}65.9 &
\cellcolor{green!6}51.2 & \cellcolor{green!6}\textbf{46.3} \\
PRP (\textsc{Gemma-2-9b-it}) &
\cellcolor{red!6}3.756 & \cellcolor{red!6}93.9 &
\cellcolor{blue!6}12.2 & \cellcolor{blue!6}80.5 & \cellcolor{blue!6}7.3 & \cellcolor{blue!6}92.7 &
\cellcolor{green!6}61.0 & \cellcolor{green!6}24.4 \\
\midrule

Greedy (\textsc{Llama-3-8B-Instruct}) &
\cellcolor{red!6}\textbf{4.0} & \cellcolor{red!6}82.9 &
\cellcolor{blue!6}36.6 & \cellcolor{blue!6}63.4 & \cellcolor{blue!6}\textbf{0.0} & \cellcolor{blue!6}\textbf{100.0} &
\cellcolor{green!6}7.3 & \cellcolor{green!6}4.9 \\
Formal (\textsc{Llama-3-8B-Instruct}) &
\cellcolor{red!6}3.341 & \cellcolor{red!6}83.5 &
\cellcolor{blue!6}\textbf{19.5} & \cellcolor{blue!6}\textbf{80.5} & \cellcolor{blue!6}\textbf{0.0} & \cellcolor{blue!6}\textbf{100.0} &
\cellcolor{green!6}0.0 & \cellcolor{green!6}0.0 \\
Global (\textsc{Llama-3-8B-Instruct}) &
\cellcolor{red!6}3.878 & \cellcolor{red!6}\textbf{97.0} &
\cellcolor{blue!6}\textbf{19.5} & \cellcolor{blue!6}56.1 & \cellcolor{blue!6}24.4 & \cellcolor{blue!6}75.6 &
\cellcolor{green!6}\textbf{31.7} & \cellcolor{green!6}9.8 \\
Weighted (\textsc{Llama-3-8B-Instruct}) &
\cellcolor{red!6}3.537 & \cellcolor{red!6}88.4 &
\cellcolor{blue!6}\textbf{19.5} & \cellcolor{blue!6}41.5 & \cellcolor{blue!6}39.0 & \cellcolor{blue!6}61.0 &
\cellcolor{green!6}17.1 & \cellcolor{green!6}\textbf{14.6} \\
PRP (\textsc{Llama-3-8B-Instruct}) &
\cellcolor{red!6}3.317 & \cellcolor{red!6}82.9 &
\cellcolor{blue!6}41.5 & \cellcolor{blue!6}51.2 & \cellcolor{blue!6}7.3 & \cellcolor{blue!6}92.7 &
\cellcolor{green!6}14.6 & \cellcolor{green!6}9.8 \\
\bottomrule
\end{tabular}%

}
\caption{Summary results with grouped metrics for S1 (one unique repair), refer Table~\ref{tab:awareus-settings}. Percentages are reported as rates (\%).}
\label{tab:oneunique}
\end{table*}


\begin{table*}[t]
\centering
\small
\renewcommand{\arraystretch}{1.18}
\setlength{\tabcolsep}{5pt}
\sisetup{
  table-number-alignment = center,
  round-mode = places,
  round-precision = 1
}
\resizebox{\textwidth}{!}{%
\begin{tabular}{l
S[table-format=1.3]
S[table-format=3.1]
S[table-format=3.1]
S[table-format=3.1]
S[table-format=3.1]
S[table-format=3.1]
S[table-format=3.1]
S[table-format=3.1]
}
\toprule
& \multicolumn{2}{c}{\cellcolor{red!15}\textbf{Dialogue Completion}} &
\multicolumn{4}{c}{\cellcolor{blue!15}\textbf{Planning Outcome}} &
\multicolumn{2}{c}{\cellcolor{green!15}\textbf{Oracle Agreement}} \\
\cmidrule(lr){2-3}\cmidrule(lr){4-7}\cmidrule(lr){8-9}
\textbf{Approach} &
\cellcolor{red!10}\textbf{Avg Cons.} &
\cellcolor{red!10}\textbf{Slot Comp.} &
\cellcolor{blue!10}\textbf{SAT no relax} &
\cellcolor{blue!10}\textbf{SAT after relax} &
\cellcolor{blue!10}\textbf{UNSAT} &
\cellcolor{blue!10}\textbf{Reco rate} &
\cellcolor{green!10}\textbf{Relax match} &
\cellcolor{green!10}\textbf{Car match} \\
\midrule

Greedy (\textsc{Qwen3-8B}) &
\cellcolor{red!6}3.800 & \cellcolor{red!6}95.0 &
\cellcolor{blue!6}20.0 & \cellcolor{blue!6}80.0 & \cellcolor{blue!6}\textbf{0.0} & \cellcolor{blue!6}\textbf{100.0} &
\cellcolor{green!6}12.5 & \cellcolor{green!6}12.5 \\
Formal (\textsc{Qwen3-8B}) &
\cellcolor{red!6}3.925 & \cellcolor{red!6}98.1 &
\cellcolor{blue!6}5.0 & \cellcolor{blue!6}\textbf{95.0} & \cellcolor{blue!6}\textbf{0.0} & \cellcolor{blue!6}\textbf{100.0} &
\cellcolor{green!6}10.0 & \cellcolor{green!6}5.0 \\
Global (\textsc{Qwen3-8B}) &
\cellcolor{red!6}3.850 & \cellcolor{red!6}96.2 &
\cellcolor{blue!6}12.5 & \cellcolor{blue!6}87.5 & \cellcolor{blue!6}\textbf{0.0} & \cellcolor{blue!6}\textbf{100.0} &
\cellcolor{green!6}40.0 & \cellcolor{green!6}25.0 \\
Weighted (\textsc{Qwen3-8B}) &
\cellcolor{red!6}\textbf{4.0} & \cellcolor{red!6}\textbf{100.0} &
\cellcolor{blue!6}\textbf{0.0} & \cellcolor{blue!6}85.0 & \cellcolor{blue!6}15.0 & \cellcolor{blue!6}85.0 &
\cellcolor{green!6}\textbf{55.0} & \cellcolor{green!6}\textbf{47.5} \\
PRP (\textsc{Qwen3-8B}) &
\cellcolor{red!6}3.675 & \cellcolor{red!6}91.9 &
\cellcolor{blue!6}30.0 & \cellcolor{blue!6}70.0 & \cellcolor{blue!6}\textbf{0.0} & \cellcolor{blue!6}\textbf{100.0} &
\cellcolor{green!6}12.5 & \cellcolor{green!6}10.0 \\
\midrule

Greedy (\textsc{Gemma-2-9b-it}) &
\cellcolor{red!6}3.9 & \cellcolor{red!6}99.4 &
\cellcolor{blue!6}15.0 & \cellcolor{blue!6}85.0 & \cellcolor{blue!6}\textbf{0.0} & \cellcolor{blue!6}\textbf{100.0} &
\cellcolor{green!6}20.0 & \cellcolor{green!6}20.0 \\
Formal (\textsc{Gemma-2-9b-it}) &
\cellcolor{red!6}3.925 & \cellcolor{red!6}98.1 &
\cellcolor{blue!6}15.0 & \cellcolor{blue!6}85.0 & \cellcolor{blue!6}\textbf{0.0} & \cellcolor{blue!6}\textbf{100.0} &
\cellcolor{green!6}0.0 & \cellcolor{green!6}0.0 \\
Global (\textsc{Gemma-2-9b-it}) &
\cellcolor{red!6}\textbf{4.0} & \cellcolor{red!6}\textbf{100.0} &
\cellcolor{blue!6}\textbf{5.0} & \cellcolor{blue!6}\textbf{95.0} & \cellcolor{blue!6}\textbf{0.0} & \cellcolor{blue!6}\textbf{100.0} &
\cellcolor{green!6}\textbf{42.5} & \cellcolor{green!6}\textbf{30.0} \\
Weighted (\textsc{Gemma-2-9b-it}) &
\cellcolor{red!6}3.9 & \cellcolor{red!6}99.4 &
\cellcolor{blue!6}\textbf{5.0} & \cellcolor{blue!6}82.5 & \cellcolor{blue!6}12.5 & \cellcolor{blue!6}87.5 &
\cellcolor{green!6}25.0 & \cellcolor{green!6}15.0 \\
PRP (\textsc{Gemma-2-9b-it}) &
\cellcolor{red!6}3.725 & \cellcolor{red!6}93.1 &
\cellcolor{blue!6}30.0 & \cellcolor{blue!6}67.5 & \cellcolor{blue!6}2.5 & \cellcolor{blue!6}97.5 &
\cellcolor{green!6}10.0 & \cellcolor{green!6}10.0 \\
\midrule

Greedy (\textsc{Llama-3-8B-Instruct}) &
\cellcolor{red!6}3.600 & \cellcolor{red!6}90.0 &
\cellcolor{blue!6}42.5 & \cellcolor{blue!6}57.5 & \cellcolor{blue!6}\textbf{0.0} & \cellcolor{blue!6}\textbf{100.0} &
\cellcolor{green!6}5.0 & \cellcolor{green!6}5.0 \\
Formal (\textsc{Llama-3-8B-Instruct}) &
\cellcolor{red!6}3.700 & \cellcolor{red!6}92.5 &
\cellcolor{blue!6}25.0 & \cellcolor{blue!6}\textbf{75.0} & \cellcolor{blue!6}\textbf{0.0} & \cellcolor{blue!6}\textbf{100.0} &
\cellcolor{green!6}0.0 & \cellcolor{green!6}0.0 \\
Global (\textsc{Llama-3-8B-Instruct}) &
\cellcolor{red!6}\textbf{3.9} & \cellcolor{red!6}\textbf{97.5} &
\cellcolor{blue!6}\textbf{17.5} & \cellcolor{blue!6}72.5 & \cellcolor{blue!6}10.0 & \cellcolor{blue!6}90.0 &
\cellcolor{green!6}\textbf{22.5} & \cellcolor{green!6}\textbf{20.0} \\
Weighted (\textsc{Llama-3-8B-Instruct}) &
\cellcolor{red!6}3.775 & \cellcolor{red!6}94.4 &
\cellcolor{blue!6}27.5 & \cellcolor{blue!6}45.0 & \cellcolor{blue!6}27.5 & \cellcolor{blue!6}72.5 &
\cellcolor{green!6}20.0 & \cellcolor{green!6}7.5 \\
PRP (\textsc{Llama-3-8B-Instruct}) &
\cellcolor{red!6}3.550 & \cellcolor{red!6}88.8 &
\cellcolor{blue!6}45.0 & \cellcolor{blue!6}45.0 & \cellcolor{blue!6}10.0 & \cellcolor{blue!6}90.0 &
\cellcolor{green!6}2.5 & \cellcolor{green!6}2.5 \\
\bottomrule
\end{tabular}%

}
\caption{Summary results with grouped metrics for S3 (any repair), refer Table~\ref{tab:awareus-settings}. Rates are reported as percentages.}
\label{tab:fourunique}
\end{table*}


\begin{table*}[!t]
\centering
\small
\renewcommand{\arraystretch}{1.18}
\setlength{\tabcolsep}{5pt}
\sisetup{
  table-number-alignment = center,
  round-mode = places,
  round-precision = 1
}
\resizebox{\textwidth}{!}{%
\begin{tabular}{l
S[table-format=1.3]
S[table-format=3.1]
S[table-format=3.1]
S[table-format=3.1]
S[table-format=3.1]
S[table-format=3.1]
S[table-format=3.1]
S[table-format=3.1]
}
\toprule
& \multicolumn{2}{c}{\cellcolor{red!15}\textbf{Dialogue Completion}} &
\multicolumn{4}{c}{\cellcolor{blue!15}\textbf{Planning Outcome}} &
\multicolumn{2}{c}{\cellcolor{green!15}\textbf{Oracle Agreement}} \\
\cmidrule(lr){2-3}\cmidrule(lr){4-7}\cmidrule(lr){8-9}
\textbf{Approach} &
\cellcolor{red!10}\textbf{Avg Cons.} &
\cellcolor{red!10}\textbf{Slot Comp.} &
\cellcolor{blue!10}\textbf{SAT no relax} &
\cellcolor{blue!10}\textbf{SAT after relax} &
\cellcolor{blue!10}\textbf{UNSAT} &
\cellcolor{blue!10}\textbf{Reco rate} &
\cellcolor{green!10}\textbf{Relax match} &
\cellcolor{green!10}\textbf{Car match} \\
\midrule

Greedy (\textsc{Qwen3-8B}) &
\cellcolor{red!6}1.610 & \cellcolor{red!6}80.5 &
\cellcolor{blue!6}34.1 & \cellcolor{blue!6}65.9 & \cellcolor{blue!6}\textbf{0.0} & \cellcolor{blue!6}\textbf{100.0} &
\cellcolor{green!6}48.8 & \cellcolor{green!6}48.8 \\
Formal (\textsc{Qwen3-8B}) &
\cellcolor{red!6}1.732 & \cellcolor{red!6}86.6 &
\cellcolor{blue!6}22.0 & \cellcolor{blue!6}78.0 & \cellcolor{blue!6}\textbf{0.0} & \cellcolor{blue!6}\textbf{100.0} &
\cellcolor{green!6}22.0 & \cellcolor{green!6}17.1 \\
Global (\textsc{Qwen3-8B}) &
\cellcolor{red!6}\textbf{1.9} & \cellcolor{red!6}\textbf{96.3} &
\cellcolor{blue!6}\textbf{7.3} & \cellcolor{blue!6}\textbf{90.2} & \cellcolor{blue!6}2.4 & \cellcolor{blue!6}97.6 &
\cellcolor{green!6}\textbf{65.9} & \cellcolor{green!6}24.4 \\
Weighted (\textsc{Qwen3-8B}) &
\cellcolor{red!6}1.756 & \cellcolor{red!6}87.8 &
\cellcolor{blue!6}22.0 & \cellcolor{blue!6}75.6 & \cellcolor{blue!6}2.4 & \cellcolor{blue!6}97.6 &
\cellcolor{green!6}63.4 & \cellcolor{green!6}\textbf{58.5} \\
PRP (\textsc{Qwen3-8B}) &
\cellcolor{red!6}1.585 & \cellcolor{red!6}79.3 &
\cellcolor{blue!6}36.6 & \cellcolor{blue!6}63.4 & \cellcolor{blue!6}\textbf{0.0} & \cellcolor{blue!6}\textbf{100.0} &
\cellcolor{green!6}48.8 & \cellcolor{green!6}22.0 \\
\midrule

Greedy (\textsc{Gemma-2-9b-it}) &
\cellcolor{red!6}1.805 & \cellcolor{red!6}90.2 &
\cellcolor{blue!6}19.5 & \cellcolor{blue!6}80.5 & \cellcolor{blue!6}\textbf{0.0} & \cellcolor{blue!6}\textbf{100.0} &
\cellcolor{green!6}51.2 & \cellcolor{green!6}48.8 \\
Formal (\textsc{Gemma-2-9b-it}) &
\cellcolor{red!6}1.878 & \cellcolor{red!6}93.9 &
\cellcolor{blue!6}12.2 & \cellcolor{blue!6}87.8 & \cellcolor{blue!6}\textbf{0.0} & \cellcolor{blue!6}\textbf{100.0} &
\cellcolor{green!6}4.9 & \cellcolor{green!6}2.4 \\
Global (\textsc{Gemma-2-9b-it}) &
\cellcolor{red!6}\textbf{2.0} & \cellcolor{red!6}\textbf{98.8} &
\cellcolor{blue!6}\textbf{4.9} & \cellcolor{blue!6}\textbf{95.1} & \cellcolor{blue!6}\textbf{0.0} & \cellcolor{blue!6}\textbf{100.0} &
\cellcolor{green!6}48.8 & \cellcolor{green!6}17.1 \\
Weighted (\textsc{Gemma-2-9b-it}) &
\cellcolor{red!6}1.854 & \cellcolor{red!6}92.7 &
\cellcolor{blue!6}14.6 & \cellcolor{blue!6}85.4 & \cellcolor{blue!6}\textbf{0.0} & \cellcolor{blue!6}\textbf{100.0} &
\cellcolor{green!6}\textbf{56.1} & \cellcolor{green!6}\textbf{53.7} \\
PRP (\textsc{Gemma-2-9b-it}) &
\cellcolor{red!6}1.780 & \cellcolor{red!6}89.0 &
\cellcolor{blue!6}24.4 & \cellcolor{blue!6}75.6 & \cellcolor{blue!6}\textbf{0.0} & \cellcolor{blue!6}\textbf{100.0} &
\cellcolor{green!6}41.5 & \cellcolor{green!6}22.0 \\
\midrule

Greedy (\textsc{Llama-3-8B-Instruct}) &
\cellcolor{red!6}1.585 & \cellcolor{red!6}79.3 &
\cellcolor{blue!6}41.5 & \cellcolor{blue!6}58.5 & \cellcolor{blue!6}\textbf{0.0} & \cellcolor{blue!6}\textbf{100.0} &
\cellcolor{green!6}29.3 & \cellcolor{green!6}26.8 \\
Formal (\textsc{Llama-3-8B-Instruct}) &
\cellcolor{red!6}1.707 & \cellcolor{red!6}85.4 &
\cellcolor{blue!6}31.7 & \cellcolor{blue!6}68.3 & \cellcolor{blue!6}\textbf{0.0} & \cellcolor{blue!6}\textbf{100.0} &
\cellcolor{green!6}12.2 & \cellcolor{green!6}9.8 \\
Global (\textsc{Llama-3-8B-Instruct}) &
\cellcolor{red!6}\textbf{1.9} & \cellcolor{red!6}\textbf{97.6} &
\cellcolor{blue!6}\textbf{4.9} & \cellcolor{blue!6}\textbf{90.2} & \cellcolor{blue!6}4.9 & \cellcolor{blue!6}95.1 &
\cellcolor{green!6}\textbf{53.7} & \cellcolor{green!6}24.4 \\
Weighted (\textsc{Llama-3-8B-Instruct}) &
\cellcolor{red!6}1.756 & \cellcolor{red!6}87.8 &
\cellcolor{blue!6}26.8 & \cellcolor{blue!6}61.0 & \cellcolor{blue!6}12.2 & \cellcolor{blue!6}87.8 &
\cellcolor{green!6}41.5 & \cellcolor{green!6}\textbf{34.1} \\
PRP (\textsc{Llama-3-8B-Instruct}) &
\cellcolor{red!6}1.610 & \cellcolor{red!6}80.5 &
\cellcolor{blue!6}41.5 & \cellcolor{blue!6}58.5 & \cellcolor{blue!6}\textbf{0.0} & \cellcolor{blue!6}\textbf{100.0} &
\cellcolor{green!6}19.5 & \cellcolor{green!6}7.3 \\
\bottomrule
\end{tabular}%

}
\caption{Summary results with grouped metrics for S2 (any unique repair), refer Table~\ref{tab:awareus-settings}. Rates are reported as percentages.}
\label{tab:twounique}
\end{table*}


\begin{figure}[t]
\centering
\begin{subfigure}[t]{0.49\columnwidth}
  \centering
  \includegraphics[width=\linewidth]{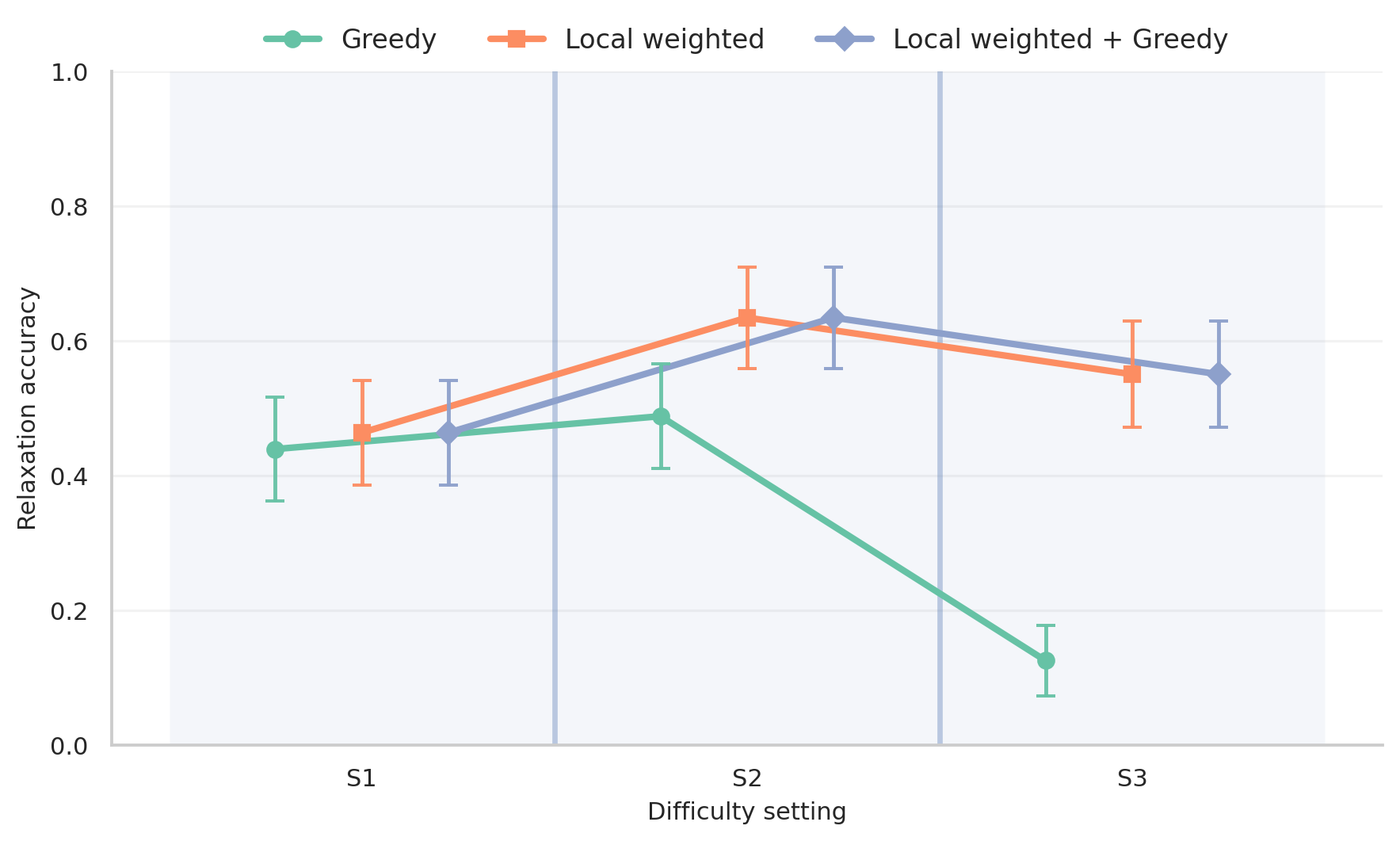}
  \caption{Relaxation accuracy across settings (S1--S3).}
  \label{fig:awareus_relax_acc}
\end{subfigure}\hfill
\begin{subfigure}[t]{0.49\columnwidth}
  \centering
  \includegraphics[width=\linewidth]{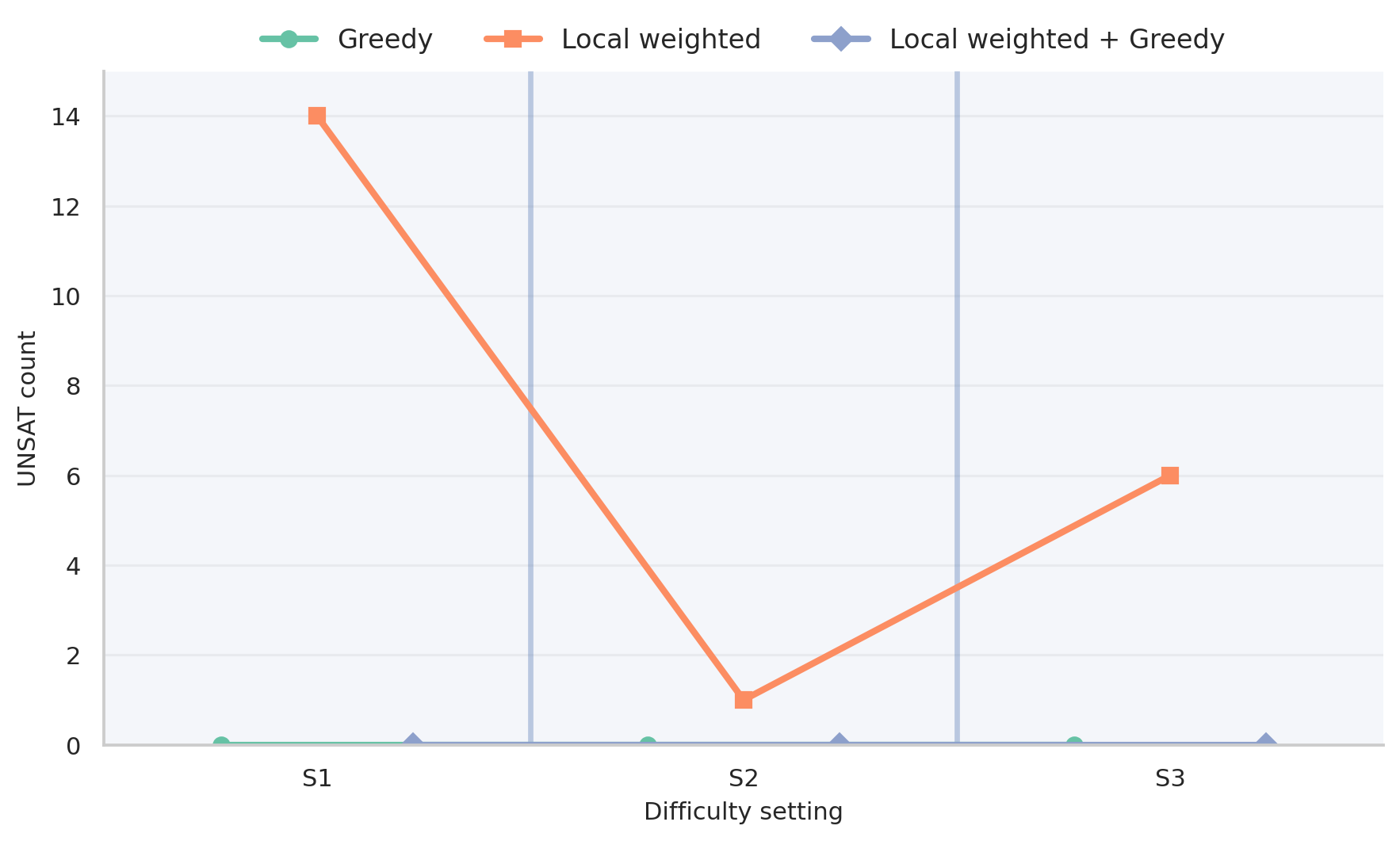}
  \caption{UNSAT count across settings (S1--S3).}
  \label{fig:awareus_unsat}
\end{subfigure}

\vspace{2pt}
\caption{\textbf{AWARE-US results by difficulty setting.} Left: relaxation accuracy. Right: UNSAT count.}
\label{fig:awareus_two_panel}
\end{figure}

For our experiments, we evaluated three large language models, Qwen3-8B \citep{yang2025qwen3technicalreport}, Gemma-2-9B-IT \citep{gemmateam2024gemma2improvingopen}, and Llama-3-8B-Instruct \citep{grattafiori2024llama3herdmodels}, across the three benchmark settings listed in Table \ref{tab:awareus-settings}. These experiments were conducted using three different frameworks as well as a baseline method; further details are provided in Section \ref{infeasibility task}. 

Tables \ref{tab:oneunique}--\ref{tab:twounique} report results for base large language models evaluated using our frameworks, with no additional training. Table \ref{tab:trained_untrained_delta_inline} compares Qwen3-8B trained using our approach (see Section \ref{training}) against local and global baselines on our test set.
\textbf{Dialogue-stage analysis.}
Across all settings, the dialogue stage is generally strong for the weighting-based agents: average extracted constraints are close to the gold count, and slot completion is typically high (mid-to-high 90s for Qwen and Gemma). Pairwise ranking (PRP) shows consistently lower slot completion and fewer extracted constraints, which already hints that some of its downstream performance loss is attributable to weaker constraint recovery, not only ranking quality.

\textbf{S1 (one unique repair).} Table \ref{tab:oneunique} is the hardest setting: exactly one of four constraints must be relaxed. The formal-verification-style baseline achieves very high feasibility and recommendation rates (often near 100\%), but its oracle agreement is poor (low relax match and especially low car match). This shows that ``minimal change'' repairs tend to prioritize what is easy to modify rather than what the user is willing to trade off. In contrast, local weighting yields the strongest preference-faithful outcomes in this strict regime, with the best car match for Qwen and Gemma. Global one-shot weighting can improve relax match, but it is less consistent at translating that into the oracle's final recommendation, suggesting that transcript-level weights can still miss localized trade-off signals.

\textbf{S2 (one unique repair).} Table \ref{tab:twounique} reduces complexity to two constraints, so performance depends more directly on modeling ``which matters more.'' Here, local weighting shows the clearest gains across all model families, achieving the highest car match for Qwen3-8B, Gemma-2-9B, and Llama-8B. This supports the interpretation that local weighting is capturing persona-conditioned importance rather than benefiting from search heuristics.

\textbf{S3 (any repair).} Table \ref{tab:fourunique} is more permissive: multiple relaxations can restore feasibility. As expected, most methods improve on feasibility. However, weighting remains advantageous for alignment: local weighting typically produces the highest car match (e.g., Qwen), indicating that even when feasibility is easy, preference signals help select the ``right'' repair and recommendation rather than an arbitrary satisfiable one. PRP remains weaker overall, which is consistent with its slightly lower slot completion and the added difficulty of turning pairwise comparisons into a stable repair-and-selection policy.

Figure~\ref{fig:awareus_two_panel} shows that in the unique-repair setting (S1), relaxation accuracy is similar across methods, suggesting the main bottleneck is upstream (constraint extraction/importance inference) rather than the repair heuristic. In multi-repair settings (S2/S3), \textsc{Greedy} drops, especially in S3, while preference-aware methods stay higher, consistent with greedy optimizing for quick satisfiability instead of relaxing low-importance constraints. The bottom panel shows \textsc{Local weighted} has the most UNSAT outcomes (notably in S1), indicating a weighting-only policy can choose a wrong relaxation and fail without recovery; adding a greedy fallback (\textsc{Local weighted+Greedy}) leaves relaxation accuracy largely unchanged but eliminates UNSAT, boosting SAT at the cost of less preference-driven choices.

\textbf{Trained Approach} We fine-tuned Qwen3-8B \citep{yang2025qwen3technicalreport} using the two-stage procedure in Section~\ref{infeasibility task} and re-evaluated \textsc{Local} and \textsc{Global} on the full test split. Training mainly mitigates two dominant errors: \emph{silent underspecification} (missing persona-backed constraints) and \emph{preference-inconsistent repair} (relaxing the wrong constraint). For \textsc{Local}, training strengthens slot/constraint extraction, yielding cleaner infeasibility checks and a less noisy constraint set for repair, which reduces dead-end repairs (lower UNSAT) and improves end-to-end alignment (higher gated Car match) by better preserving what is negotiable vs.\ fixed. For \textsc{Global}, slot recovery remains the bottleneck, but relaxations become more oracle-consistent (higher Relax match); Car match improves less, likely because single-pass global weighting is sensitive to parsing noise and can blur fine-grained trade-offs. Overall, training helps both, with the largest gains when preferences are grounded in slot-level evidence (\textsc{Local}).

\textbf{Discussion}
Performance is limited by two bottlenecks: (i) recovering persona-backed constraints through dialogue and (ii) selecting the oracle-consistent relaxation when infeasible. Training helps mainly by reducing \emph{silent underspecification}, producing cleaner constraint sets that improve SAT-after-relaxation and reduce UNSAT.

Weighted relaxation is the most reliable way to resolve infeasibility while preserving preferences: gains in Relax match track gains in gated Car match, indicating the model is not just restoring feasibility but relaxing the right constraint. In contrast, minimal-change baselines often achieve high feasibility yet lag in oracle agreement because “smallest edit” need not be the least important edit. Local vs.\ global weighting reflects a trade-off: global weighting can better infer importance from multi-turn context but is more sensitive to parsing noise and may lose fine-grained intent for final ranking; local weighting ties weights to slot-level evidence and benefits more from improved extraction, yielding stronger end-to-end agreement after training.

\newcommand{\Up}[1]{\textcolor{blue}{\(\uparrow\)\,#1}}
\newcommand{\Down}[1]{\textcolor{red}{\(\downarrow\)\,#1}}
\newcommand{\Flat}[1]{\textcolor{black}{#1}}

\newcommand{\inc}[1]{\textcolor{blue}{\scriptsize\(\uparrow\)+#1\%\,}}
\newcommand{\dec}[1]{\textcolor{red}{\scriptsize\(\downarrow\)#1\%\,}}

\setlength{\textfloatsep}{4pt}
\setlength{\intextsep}{4pt}
\begin{table}[t]
\centering
\small
\renewcommand{\arraystretch}{1.10}
\setlength{\tabcolsep}{3pt}
\resizebox{\columnwidth}{!}{%
\begin{tabular}{lrrrrrrrr}
\toprule
& \multicolumn{2}{c}{\cellcolor{red!15}\textbf{Dialogue Completion}} &
\multicolumn{4}{c}{\cellcolor{blue!15}\textbf{Planning Outcome}} &
\multicolumn{2}{c}{\cellcolor{green!15}\textbf{Oracle Agreement}} \\
\cmidrule(lr){2-3}\cmidrule(lr){4-7}\cmidrule(lr){8-9}
\textbf{Setting} &
\cellcolor{red!10}\textbf{Avg slots} &
\cellcolor{red!10}\textbf{Slot comp.} &
\cellcolor{blue!10}\textbf{SAT no relax} &
\cellcolor{blue!10}\textbf{SAT after relax} &
\cellcolor{blue!10}\textbf{UNSAT} &
\cellcolor{blue!10}\textbf{Reco rate} &
\cellcolor{green!10}\textbf{Relax match} &
\cellcolor{green!10}\textbf{Car match} \\
\midrule
Local (Base.)  &
\cellcolor{red!6}3.28 &
\cellcolor{red!6}83.9 &
\cellcolor{blue!6}12.0 &
\cellcolor{blue!6}68.0 &
\cellcolor{blue!6}20.0 &
\cellcolor{blue!6}80.0 &
\cellcolor{green!6}44.0 &
\cellcolor{green!6}36.0 \\
Local (Tr.)    &
\cellcolor{red!6}3.28 &
\cellcolor{red!6}\inc{16.1}\textbf{100.0} &
\cellcolor{blue!6}\dec{12.0}0.0 &
\cellcolor{blue!6}\inc{20.0}\textbf{88.0} &
\cellcolor{blue!6}\dec{8.0}12.0 &
\cellcolor{blue!6}\inc{8.0}88.0 &
\cellcolor{green!6}\inc{8.0}52.0 &
\cellcolor{green!6}\inc{12.0}\textbf{48.0} \\
\midrule
Global (Base.) &
\cellcolor{red!6}3.28 &
\cellcolor{red!6}90.2 &
\cellcolor{blue!6}20.0 &
\cellcolor{blue!6}72.0 &
\cellcolor{blue!6}4.0 &
\cellcolor{blue!6}\textbf{96.0} &
\cellcolor{green!6}40.0 &
\cellcolor{green!6}28.0 \\
Global (Tr.)   &
\cellcolor{red!6}3.28 &
\cellcolor{red!6}90.2 &
\cellcolor{blue!6}\inc{4.0}24.0 &
\cellcolor{blue!6}\inc{4.0}76.0 &
\cellcolor{blue!6}4.0 &
\cellcolor{blue!6}\textbf{96.0} &
\cellcolor{green!6}\inc{16.0}\textbf{56.0} &
\cellcolor{green!6}\inc{8.0}36.0 \\
\bottomrule
\end{tabular}%
}
\caption{Trained vs Base Qwen3-8B under Local and Global settings on the test set. Values are rates (no \%). Trained rows show $\Delta$ vs Base as a left prefix: blue \inc{k} improves, red \dec{k} degrades. \textbf{Car match} is gated on correct relaxation.}
\label{tab:trained_untrained_delta_inline}
\end{table}

S1 settings show infeasibility resolution spans regimes: when many relaxations work, methods look similar on feasibility and oracle agreement differentiates; when only one relaxation is valid, the task becomes sharply preference-sensitive, widening the gap between preference-aware weighting and heuristic/minimal-change methods.
\section{Limitations and Future Work.}

Our study is constrained by computational limits, which restricted training scale, hyperparameter sweeps, and the range of model sizes evaluated. Benchmark-wise, \textsc{AWARE-US} currently targets a structured car domain with a finite constraint vocabulary and persona-conditioned supervision (oracle repairs/weights), which may not capture the diversity, noise, domain shift, and preference drift found in real interactions, especially in settings like travel planning \citep{xie2024travelplannerbenchmarkrealworldplanning}; moreover, our framework assumes largely \emph{static} preferences and does not explicitly handle time-varying priorities or mid-dialogue preference changes. A natural next step is to expand \textsc{AWARE-US} in size and domain coverage with richer constraints and more challenging infeasibility patterns, and to strengthen optimization by exploring improved preference-pair generation , trajectory-level preference optimization that scores the full repair-and-recommend sequence \citep{zhang2024askbeforeplanproactivelanguageagents}, and hybrid recipes that combine supervised learning for robust preference extraction with DPO-style objectives to better preserve fine-grained trade-offs during final selection.

\section{Conclusion.}
This paper introduced \emph{Preference-Aware Resolution} for tool-calling agents: when a fully specified database query becomes infeasible, the agent should relax the \emph{least important} constraint using preference cues from dialogue, rather than fixed heuristics or trial-and-error. We proposed three LLM-based importance-inference methods (Section \ref{infeasibility task}) and released AWARE-US, a persona-grounded benchmark that couples underspecification handling with preference-faithful infeasibility repair and recommendation selection. Results across models and settings show weighting-based strategies especially local weighting, yielding more intent-consistent repairs and higher agreement with oracle preferences than minimal-change baselines, underscoring the value of explicitly modeling user-specific constraint importance.

\section*{Acknowledgements}
I thank Prof.\ Alexandra Birch for her valuable guidance, insightful feedback, and helpful discussions throughout this project. This work was supported by \textbf{Unified Transcription and Translation for Extended Reality (UTTER)}.

\bibliography{colm2026_conference}

@misc{wang2025surveyevolutionlanguagemodelbased,
      title={A Survey of the Evolution of Language Model-Based Dialogue Systems: Data, Task and Models}, 
      author={Hongru Wang and Lingzhi Wang and Yiming Du and Liang Chen and Jingyan Zhou and Yufei Wang and Kam-Fai Wong},
      year={2025},
      eprint={2311.16789},
      archivePrefix={arXiv},
      primaryClass={cs.CL},
      url={https://arxiv.org/abs/2311.16789}, 
}

@article{acikgoz2024convagents,
  title   = "The Rise of Conversational AI Agents with Large Language Models",
  author  = "Emre Can Acikgoz and Dilek Hakkani-Tur and Gokhan Tur",
  journal = "emrecanacikgoz.github.io",
  year    = "2024",
  month   = "November",
  url     = "https://emrecanacikgoz.github.io/Conversational-Agents/"
}

@misc{qin2023endtoendtaskorienteddialoguesurvey,
      title={End-to-end Task-oriented Dialogue: A Survey of Tasks, Methods, and Future Directions}, 
      author={Libo Qin and Wenbo Pan and Qiguang Chen and Lizi Liao and Zhou Yu and Yue Zhang and Wanxiang Che and Min Li},
      year={2023},
      eprint={2311.09008},
      archivePrefix={arXiv},
      primaryClass={cs.CL},
      url={https://arxiv.org/abs/2311.09008}, 
}

@misc{zhang2024askbeforeplanproactivelanguageagents,
      title={Ask-before-Plan: Proactive Language Agents for Real-World Planning}, 
      author={Xuan Zhang and Yang Deng and Zifeng Ren and See-Kiong Ng and Tat-Seng Chua},
      year={2024},
      eprint={2406.12639},
      archivePrefix={arXiv},
      primaryClass={cs.CL},
      url={https://arxiv.org/abs/2406.12639}, 
}

@misc{hao2025largelanguagemodelssolve,
      title={Large Language Models Can Solve Real-World Planning Rigorously with Formal Verification Tools}, 
      author={Yilun Hao and Yongchao Chen and Yang Zhang and Chuchu Fan},
      year={2025},
      eprint={2404.11891},
      archivePrefix={arXiv},
      primaryClass={cs.AI},
      url={https://arxiv.org/abs/2404.11891}, 
}

@misc{xie2024travelplannerbenchmarkrealworldplanning,
      title={TravelPlanner: A Benchmark for Real-World Planning with Language Agents}, 
      author={Jian Xie and Kai Zhang and Jiangjie Chen and Tinghui Zhu and Renze Lou and Yuandong Tian and Yanghua Xiao and Yu Su},
      year={2024},
      eprint={2402.01622},
      archivePrefix={arXiv},
      primaryClass={cs.CL},
      url={https://arxiv.org/abs/2402.01622}, 
}

@inproceedings{
anonymous2025argus,
title={Argus: Disambiguating User Queries for Tool-Calling Agents via Uncertainty Quantification},
author={Anonymous},
booktitle={Submitted to ACL Rolling Review - May 2025},
year={2025},
url={https://openreview.net/forum?id=U1SpjHP8gh},
note={under review}
}

@misc{shim2025tooldialmultiturndialoguegeneration,
      title={ToolDial: Multi-turn Dialogue Generation Method for Tool-Augmented Language Models}, 
      author={Jeonghoon Shim and Gyuhyeon Seo and Cheongsu Lim and Yohan Jo},
      year={2025},
      eprint={2503.00564},
      archivePrefix={arXiv},
      primaryClass={cs.CL},
      url={https://arxiv.org/abs/2503.00564}, 
}

@misc{xi2023risepotentiallargelanguage,
      title={The Rise and Potential of Large Language Model Based Agents: A Survey}, 
      author={Zhiheng Xi and Wenxiang Chen and Xin Guo and Wei He and Yiwen Ding and Boyang Hong and Ming Zhang and Junzhe Wang and Senjie Jin and Enyu Zhou and Rui Zheng and Xiaoran Fan and Xiao Wang and Limao Xiong and Yuhao Zhou and Weiran Wang and Changhao Jiang and Yicheng Zou and Xiangyang Liu and Zhangyue Yin and Shihan Dou and Rongxiang Weng and Wensen Cheng and Qi Zhang and Wenjuan Qin and Yongyan Zheng and Xipeng Qiu and Xuanjing Huang and Tao Gui},
      year={2023},
      eprint={2309.07864},
      archivePrefix={arXiv},
      primaryClass={cs.AI},
      url={https://arxiv.org/abs/2309.07864}, 
}

@article{Wang_2024,
   title={A survey on large language model based autonomous agents},
   volume={18},
   ISSN={2095-2236},
   url={http://dx.doi.org/10.1007/s11704-024-40231-1},
   DOI={10.1007/s11704-024-40231-1},
   number={6},
   journal={Frontiers of Computer Science},
   publisher={Springer Science and Business Media LLC},
   author={Wang, Lei and Ma, Chen and Feng, Xueyang and Zhang, Zeyu and Yang, Hao and Zhang, Jingsen and Chen, Zhiyuan and Tang, Jiakai and Chen, Xu and Lin, Yankai and Zhao, Wayne Xin and Wei, Zhewei and Wen, Jirong},
   year={2024},
   month=mar }

@inproceedings{deng2024large,
  title={Large language model powered agents in the web},
  author={Deng, Yang and Zhang, An and Lin, Yankai and Chen, Xu and Wen, Ji-Rong and Chua, Tat-Seng},
  booktitle={Companion Proceedings of the ACM Web Conference 2024},
  pages={1242--1245},
  year={2024}
}

@inproceedings{yao2022react,
  title={React: Synergizing reasoning and acting in language models},
  author={Yao, Shunyu and Zhao, Jeffrey and Yu, Dian and Du, Nan and Shafran, Izhak and Narasimhan, Karthik R and Cao, Yuan},
  booktitle={The eleventh international conference on learning representations},
  year={2022}
}

@misc{shinn2023reflexionlanguageagentsverbal,
      title={Reflexion: Language Agents with Verbal Reinforcement Learning}, 
      author={Noah Shinn and Federico Cassano and Edward Berman and Ashwin Gopinath and Karthik Narasimhan and Shunyu Yao},
      year={2023},
      eprint={2303.11366},
      archivePrefix={arXiv},
      primaryClass={cs.AI},
      url={https://arxiv.org/abs/2303.11366}, 
}

@misc{qin2023toolllmfacilitatinglargelanguage,
      title={ToolLLM: Facilitating Large Language Models to Master 16000+ Real-world APIs}, 
      author={Yujia Qin and Shihao Liang and Yining Ye and Kunlun Zhu and Lan Yan and Yaxi Lu and Yankai Lin and Xin Cong and Xiangru Tang and Bill Qian and Sihan Zhao and Lauren Hong and Runchu Tian and Ruobing Xie and Jie Zhou and Mark Gerstein and Dahai Li and Zhiyuan Liu and Maosong Sun},
      year={2023},
      eprint={2307.16789},
      archivePrefix={arXiv},
      primaryClass={cs.AI},
      url={https://arxiv.org/abs/2307.16789}, 
}

@misc{wen2024diluknowledgedrivenapproachautonomous,
      title={DiLu: A Knowledge-Driven Approach to Autonomous Driving with Large Language Models}, 
      author={Licheng Wen and Daocheng Fu and Xin Li and Xinyu Cai and Tao Ma and Pinlong Cai and Min Dou and Botian Shi and Liang He and Yu Qiao},
      year={2024},
      eprint={2309.16292},
      archivePrefix={arXiv},
      primaryClass={cs.RO},
      url={https://arxiv.org/abs/2309.16292}, 
}

@misc{yao2023reactsynergizingreasoningacting,
      title={ReAct: Synergizing Reasoning and Acting in Language Models}, 
      author={Shunyu Yao and Jeffrey Zhao and Dian Yu and Nan Du and Izhak Shafran and Karthik Narasimhan and Yuan Cao},
      year={2023},
      eprint={2210.03629},
      archivePrefix={arXiv},
      primaryClass={cs.CL},
      url={https://arxiv.org/abs/2210.03629}, 
}

@misc{schick2023toolformerlanguagemodelsteach,
      title={Toolformer: Language Models Can Teach Themselves to Use Tools}, 
      author={Timo Schick and Jane Dwivedi-Yu and Roberto Dessì and Roberta Raileanu and Maria Lomeli and Luke Zettlemoyer and Nicola Cancedda and Thomas Scialom},
      year={2023},
      eprint={2302.04761},
      archivePrefix={arXiv},
      primaryClass={cs.CL},
      url={https://arxiv.org/abs/2302.04761}, 
}

@misc{patil2023gorillalargelanguagemodel,
      title={Gorilla: Large Language Model Connected with Massive APIs}, 
      author={Shishir G. Patil and Tianjun Zhang and Xin Wang and Joseph E. Gonzalez},
      year={2023},
      eprint={2305.15334},
      archivePrefix={arXiv},
      primaryClass={cs.CL},
      url={https://arxiv.org/abs/2305.15334}, 
}

@misc{li2023apibankcomprehensivebenchmarktoolaugmented,
      title={API-Bank: A Comprehensive Benchmark for Tool-Augmented LLMs}, 
      author={Minghao Li and Yingxiu Zhao and Bowen Yu and Feifan Song and Hangyu Li and Haiyang Yu and Zhoujun Li and Fei Huang and Yongbin Li},
      year={2023},
      eprint={2304.08244},
      archivePrefix={arXiv},
      primaryClass={cs.CL},
      url={https://arxiv.org/abs/2304.08244}, 
}

@misc{andukuri2024stargateteachinglanguagemodels,
      title={STaR-GATE: Teaching Language Models to Ask Clarifying Questions}, 
      author={Chinmaya Andukuri and Jan-Philipp Fränken and Tobias Gerstenberg and Noah D. Goodman},
      year={2024},
      eprint={2403.19154},
      archivePrefix={arXiv},
      primaryClass={cs.CL},
      url={https://arxiv.org/abs/2403.19154}, 
}

@misc{yu2019sparccrossdomainsemanticparsing,
      title={SParC: Cross-Domain Semantic Parsing in Context}, 
      author={Tao Yu and Rui Zhang and Michihiro Yasunaga and Yi Chern Tan and Xi Victoria Lin and Suyi Li and Heyang Er and Irene Li and Bo Pang and Tao Chen and Emily Ji and Shreya Dixit and David Proctor and Sungrok Shim and Jonathan Kraft and Vincent Zhang and Caiming Xiong and Richard Socher and Dragomir Radev},
      year={2019},
      eprint={1906.02285},
      archivePrefix={arXiv},
      primaryClass={cs.CL},
      url={https://arxiv.org/abs/1906.02285}, 
}

@misc{yu2019cosqlconversationaltexttosqlchallenge,
      title={CoSQL: A Conversational Text-to-SQL Challenge Towards Cross-Domain Natural Language Interfaces to Databases}, 
      author={Tao Yu and Rui Zhang and He Yang Er and Suyi Li and Eric Xue and Bo Pang and Xi Victoria Lin and Yi Chern Tan and Tianze Shi and Zihan Li and Youxuan Jiang and Michihiro Yasunaga and Sungrok Shim and Tao Chen and Alexander Fabbri and Zifan Li and Luyao Chen and Yuwen Zhang and Shreya Dixit and Vincent Zhang and Caiming Xiong and Richard Socher and Walter S Lasecki and Dragomir Radev},
      year={2019},
      eprint={1909.05378},
      archivePrefix={arXiv},
      primaryClass={cs.CL},
      url={https://arxiv.org/abs/1909.05378}, 
}

@article{Jannach_2021,
   title={A Survey on Conversational Recommender Systems},
   volume={54},
   ISSN={1557-7341},
   url={http://dx.doi.org/10.1145/3453154},
   DOI={10.1145/3453154},
   number={5},
   journal={ACM Computing Surveys},
   publisher={Association for Computing Machinery (ACM)},
   author={Jannach, Dietmar and Manzoor, Ahtsham and Cai, Wanling and Chen, Li},
   year={2021},
   month=may, pages={1–36} }

@inproceedings{ikeda2024efficient,
  title={An Efficient Diversity-Aware Method for the Empty-Answer Problem.},
  author={Ikeda, Yuto and Xiao, Chuan and Onizuka, Makoto},
  booktitle={DOLAP},
  pages={68--72},
  year={2024}
}

@inproceedings{mottin2014iqr,
  title={IQR: an interactive query relaxation system for the empty-answer problem},
  author={Mottin, Davide and Marascu, Alice and Basu Roy, Senjuti and Das, Gautam and Palpanas, Themis and Velegrakis, Yannis},
  booktitle={Proceedings of the 2014 ACM SIGMOD international conference on Management of Data},
  pages={1095--1098},
  year={2014}
}

@article{mottin2013probabilistic,
  title={A probabilistic optimization framework for the empty-answer problem},
  author={Mottin, Davide and Marascu, Alice and Roy, Senjuti Basu and Das, Gautam and Palpanas, Themis and Velegrakis, Yannis},
  journal={Proceedings of the VLDB Endowment},
  volume={6},
  number={14},
  pages={1762--1773},
  year={2013},
  publisher={VLDB Endowment}
}

@article{oruche2025survey,
  title={A Survey on the Recent Advancements in Human-Centered Dialog Systems},
  author={Oruche, Roland and Goruganthu, Sai Keerthana and Akula, Rithika and Cheng, Xiyao and Goni, Ashraful Md and Shibo, Bruce W and Kee, Kerk and Zampieri, Marcos and Calyam, Prasad},
  journal={ACM Computing Surveys},
  volume={57},
  number={10},
  pages={1--36},
  year={2025},
  publisher={ACM New York, NY}
}

@article{chen2017survey,
  title={A survey on dialogue systems: Recent advances and new frontiers},
  author={Chen, Hongshen and Liu, Xiaorui and Yin, Dawei and Tang, Jiliang},
  journal={Acm Sigkdd Explorations Newsletter},
  volume={19},
  number={2},
  pages={25--35},
  year={2017},
  publisher={Acm New York, NY, USA}
}

@article{wang2021task,
  title={Task-oriented dialogue system as natural language generation. arxiv 2021},
  author={Wang, W and Zhang, Z and Guo, J and Dai, Y and Chen, B and Luo, W},
  journal={arXiv preprint arXiv:2108.13679},
  year={2021}
}

@article{hosseini2020simple,
  title={A simple language model for task-oriented dialogue},
  author={Hosseini-Asl, Ehsan and McCann, Bryan and Wu, Chien-Sheng and Yavuz, Semih and Socher, Richard},
  journal={Advances in Neural Information Processing Systems},
  volume={33},
  pages={20179--20191},
  year={2020}
}

@misc{ouyang2022traininglanguagemodelsfollow,
      title={Training language models to follow instructions with human feedback}, 
      author={Long Ouyang and Jeff Wu and Xu Jiang and Diogo Almeida and Carroll L. Wainwright and Pamela Mishkin and Chong Zhang and Sandhini Agarwal and Katarina Slama and Alex Ray and John Schulman and Jacob Hilton and Fraser Kelton and Luke Miller and Maddie Simens and Amanda Askell and Peter Welinder and Paul Christiano and Jan Leike and Ryan Lowe},
      year={2022},
      eprint={2203.02155},
      archivePrefix={arXiv},
      primaryClass={cs.CL},
      url={https://arxiv.org/abs/2203.02155}, 
}

@misc{rafailov2024directpreferenceoptimizationlanguage,
      title={Direct Preference Optimization: Your Language Model is Secretly a Reward Model}, 
      author={Rafael Rafailov and Archit Sharma and Eric Mitchell and Stefano Ermon and Christopher D. Manning and Chelsea Finn},
      year={2024},
      eprint={2305.18290},
      archivePrefix={arXiv},
      primaryClass={cs.LG},
      url={https://arxiv.org/abs/2305.18290}, 
}

@misc{zhang2023clarifynecessaryresolvingambiguity,
      title={Clarify When Necessary: Resolving Ambiguity Through Interaction with LMs}, 
      author={Michael J. Q. Zhang and Eunsol Choi},
      year={2023},
      eprint={2311.09469},
      archivePrefix={arXiv},
      primaryClass={cs.CL},
      url={https://arxiv.org/abs/2311.09469}, 
}

@article{liffiton2008algorithms,
  title={Algorithms for computing minimal unsatisfiable subsets of constraints},
  author={Liffiton, Mark H and Sakallah, Karem A},
  journal={Journal of Automated Reasoning},
  volume={40},
  number={1},
  pages={1--33},
  year={2008},
  publisher={Springer}
}

@inproceedings{bacchus2015using,
  title={Using minimal correction sets to more efficiently compute minimal unsatisfiable sets},
  author={Bacchus, Fahiem and Katsirelos, George},
  booktitle={International Conference on Computer Aided Verification},
  pages={70--86},
  year={2015},
  organization={Springer}
}

@article{reiter1987theory,
  title={A theory of diagnosis from first principles},
  author={Reiter, Raymond},
  journal={Artificial intelligence},
  volume={32},
  number={1},
  pages={57--95},
  year={1987},
  publisher={Elsevier}
}

@inproceedings{jandaghi2024faithful,
  title={Faithful persona-based conversational dataset generation with large language models},
  author={Jandaghi, Pegah and Sheng, XiangHai and Bai, Xinyi and Pujara, Jay and Sidahmed, Hakim},
  booktitle={Proceedings of the 6th Workshop on NLP for Conversational AI (NLP4ConvAI 2024)},
  pages={114--139},
  year={2024}
}

@misc{openai_gpt5_docs,
  title        = {GPT-5 Documentation},
  author       = {{OpenAI}},
  howpublished = {\url{https://openai.com/index/introducing-gpt-5/}},
  year         = {2025},
  note         = {Accessed: 2026-01-04}
}

@inproceedings{koudas2006relaxing,
  title={Relaxing join and selection queries},
  author={Koudas, Nick and Li, Chen and Tung, Anthony KH and Vernica, Rares},
  booktitle={Proceedings of the 32nd international conference on Very large data bases},
  pages={199--210},
  year={2006}
}

@misc{qin2024largelanguagemodelseffective,
      title={Large Language Models are Effective Text Rankers with Pairwise Ranking Prompting}, 
      author={Zhen Qin and Rolf Jagerman and Kai Hui and Honglei Zhuang and Junru Wu and Le Yan and Jiaming Shen and Tianqi Liu and Jialu Liu and Donald Metzler and Xuanhui Wang and Michael Bendersky},
      year={2024},
      eprint={2306.17563},
      archivePrefix={arXiv},
      primaryClass={cs.IR},
      url={https://arxiv.org/abs/2306.17563}, 
}

@misc{yang2025qwen3technicalreport,
      title={Qwen3 Technical Report}, 
      author={An Yang and Anfeng Li and Baosong Yang and Beichen Zhang and Binyuan Hui and Bo Zheng and Bowen Yu and Chang Gao and Chengen Huang},
      year={2025},
      eprint={2505.09388},
      archivePrefix={arXiv},
      primaryClass={cs.CL},
      url={https://arxiv.org/abs/2505.09388}, 
}

@misc{grattafiori2024llama3herdmodels,
      title={The Llama 3 Herd of Models}, 
      author={Aaron Grattafiori and Yuchen Hao and Yundi Qian and Yunlu Li and Yuzi He and Zach Rait and Zachary DeVito and Zef Rosnbrick and Zhaoduo Wen and Zhenyu Yang and Zhiwei Zhao and Zhiyu Ma},
      year={2024},
      eprint={2407.21783},
      archivePrefix={arXiv},
      primaryClass={cs.AI},
      url={https://arxiv.org/abs/2407.21783}, 
}

@misc{gemmateam2024gemma2improvingopen,
      title={Gemma 2: Improving Open Language Models at a Practical Size}, 
      author={Gemma Team and Morgane Riviere and Shreya Pathak and Pier Giuseppe Sessa and Cassidy Hardin and Surya Bhupatiraju and Léonard Hussenot and Thomas Mesnard and Bobak Shahriari and Alexandre Ramé},
      year={2024},
      eprint={2408.00118},
      archivePrefix={arXiv},
      primaryClass={cs.CL},
      url={https://arxiv.org/abs/2408.00118}, 
}

@misc{dettmers2023qloraefficientfinetuningquantized,
      title={QLoRA: Efficient Finetuning of Quantized LLMs}, 
      author={Tim Dettmers and Artidoro Pagnoni and Ari Holtzman and Luke Zettlemoyer},
      year={2023},
      eprint={2305.14314},
      archivePrefix={arXiv},
      primaryClass={cs.LG},
      url={https://arxiv.org/abs/2305.14314}, 
}

@article{tunstall2023zephyr,
  title={Zephyr: Direct distillation of lm alignment},
  author={Tunstall, Lewis and Beeching, Edward and Lambert, Nathan and Rajani, Nazneen and Rasul, Kashif and Belkada, Younes and Huang, Shengyi and Von Werra, Leandro and Fourrier, Cl{\'e}mentine and Habib, Nathan and others},
  journal={arXiv preprint arXiv:2310.16944},
  year={2023}
}

@misc{cooperunion_car_features_msrp_2016,
  author       = {CooperUnion},
  title        = {Car Features and MSRP},
  howpublished = {Kaggle dataset},
  year         = {2016},
  month        = dec,
  note         = {Published 21 Dec 2016. Accessed 4 Jan 2026},
  url          = {https://www.kaggle.com/datasets/CooperUnion/cardataset}
}

@article{corella2009brief,
  title={A brief overview of cooperative answering},
  author={Corella, F and Lewison, K},
  journal={Journal of Intelligent Information Systems},
  volume={1},
  number={2},
  pages={123--157},
  year={2009}
}
\bibliographystyle{colm2026_conference}

\clearpage
\FloatBarrier

\appendix
\section{Appendix}
\subsection{Algorithms}
\label{Algos}

\FloatBarrier
\begin{algorithm}[H]
\caption{DPO triple generation for relaxation-choice training}
\label{alg:dpo_relax_gen}
\begin{algorithmic}[1]
\Require Benchmark items $\mathcal{B}=\{(q_i,u_i,A_i,a_i^\star)\}_{i=1}^n$, database $D$, roleplayer $\mathrm{RP}$, max negatives $K$
\Ensure DPO triples $\mathcal{D}_{\mathrm{DPO}}$

\Statex \textbf{Feasibility set:}\;
$S(q,C)\triangleq\{\,r\in D:\ r \models \mathrm{base}(q)\ \wedge\ \bigwedge_{a\in C} a\,\}$.
\Statex \textbf{Canonical action:}\; $g(a)=$ JSON encoding of \texttt{relaxed\_constraint = a}.
\Statex \textbf{Prompt builder:}\; $\Phi(q,\mathcal{D},A)$ formats $(q,\mathcal{D},A)$ into the agent prompt.

\State $\mathcal{D}_{\mathrm{DPO}} \gets \emptyset$
\For{$i \gets 1$ to $n$}
  \State $(q,u,A,a^\star) \gets (q_i,u_i,A_i,a_i^\star)$
  \If{$a^\star \notin A$}
    \State \textbf{continue}
  \EndIf
  \If{$S(q,A)\neq \emptyset$} \Comment{keep only infeasible full queries}
    \State \textbf{continue}
  \EndIf

  \State $\mathcal{D} \gets \{(\mathrm{slot}(a),\,\mathrm{RP}(q,u,\mathrm{slot}(a)))\}_{a\in A}$
  \State $x \gets \Phi(q,\mathcal{D},A)$
  \State $y^{+} \gets g(a^\star)$

  \ForAll{$c \in A\setminus\{a^\star\}$}
    \State $R(c) \gets S(q, A\setminus\{c\})$
    \State $\mathrm{score}(c) \gets \big(\mathbb{I}[|R(c)|>0],\,|R(c)|\big)$
    \Comment{hard-first, then smaller $|R|$}
  \EndFor

  \State $N \gets \operatorname{TopK}\big(\{c:\mathrm{score}(c)\},\,K\big)$
  \ForAll{$c \in N$}
    \State $y^{-} \gets g(c)$
    \State $\mathcal{D}_{\mathrm{DPO}} \gets \mathcal{D}_{\mathrm{DPO}} \cup \{(x,y^{+},y^{-})\}$
  \EndFor
\EndFor
\State \Return $\mathcal{D}_{\mathrm{DPO}}$
\end{algorithmic}
\end{algorithm}
\FloatBarrier

\makeatletter
\newenvironment{breakablealgorithm}
  {%
    \begin{center}
      \refstepcounter{algorithm}%
      \hrule height .8pt depth 0pt \kern 2pt%
      \renewcommand{\caption}[2][\relax]{%
        {\raggedright\textbf{Algorithm~\thealgorithm:} ##2\par}%
        \ifx\relax##1\relax
          \addcontentsline{loa}{algorithm}{\protect\numberline{\thealgorithm}##2}%
        \else
          \addcontentsline{loa}{algorithm}{\protect\numberline{\thealgorithm}##1}%
        \fi
        \kern 2pt\hrule\kern 2pt%
      }%
  }{%
      \kern 2pt\hrule\relax%
    \end{center}
  }
\makeatother

\begin{breakablealgorithm}
\caption{Two-stage training (SFT roleplayer, then DPO agent)}
\label{alg:two_stage_training}
\footnotesize
\begin{algorithmic}[1]
\State $B_{\text{unique}} \gets \textsc{Deduplicate}(B,\ \text{key}=(q,u))$
\State $(B_{\text{train}},B_{\text{test}}) \gets \textsc{SeededSplit}(B_{\text{unique}},\ 80/20)$

\Statex \textbf{Stage 1: SFT (train ROLEPLAYER)}
\State $D_{\text{SFT}} \gets \emptyset$
\ForAll{$b \in B_{\text{train}}$}
  \State $(q,u,A,W) \gets (b.\texttt{base\_query\_sentence},\ b.\texttt{persona},\ b.\texttt{additional\_constraints},\ b.\texttt{constraint\_weights})$
  \State $S_c \gets \textsc{MapSlots}(A)$
  \State $W_c \gets \textsc{MapWeights}(W)$
  \ForAll{$slot \in \textsc{Keys}(S_c)$}
    \State $y \gets \textsc{RoleplayerTargetJSON}(slot,S_c,W_c)$
    \State $x \gets \textsc{RoleplayerPrompt}(q,u,slot)$
    \State $D_{\text{SFT}} \gets D_{\text{SFT}} \cup \{(x,y)\}$
  \EndFor
  \State $A_{\text{abs}} \gets \textsc{SLOTS} \setminus \textsc{Keys}(S_c)$
  \ForAll{$slot \in \textsc{Sample}(A_{\text{abs}},\ n=\textit{small\_constant})$}
    \State $y \gets \textsc{JSON}(\texttt{slot}=slot,\ \texttt{has\_preference}=false,\ \texttt{op}=\texttt{"=="},\ \texttt{value}=\texttt{null},\ \texttt{importance\_label}=\texttt{"NO\_PREFERENCE"})$
    \State $x \gets \textsc{RoleplayerPrompt}(q,u,slot)$
    \State $D_{\text{SFT}} \gets D_{\text{SFT}} \cup \{(x,y)\}$
  \EndFor
\EndFor
\State $\theta_{\text{RP}} \gets \textsc{SFT\_QLoRA}(M_0,\ D_{\text{SFT}},\ \text{loss}=\text{NLL on completion tokens only})$
\State \textsc{SaveAdapter}$(\theta_{\text{RP}})$
\State $\text{RP} \gets \textsc{LoadModel}(M_0,\ \text{adapter}=\theta_{\text{RP}})$

\Statex \textbf{Stage 2: DPO (train AGENT)}
\State $D_{\text{DPO}} \gets \emptyset$
\ForAll{$b \in B_{\text{train}}$}
  \State $(q,u,A,a^\star) \gets (b.\texttt{base\_query\_sentence},\ b.\texttt{persona},\ b.\texttt{additional\_constraints},\ b.\texttt{oracle\_relaxation})$
  \If{$\neg\,\textsc{IsMember}(a^\star,A)$} \State \textbf{continue} \EndIf
  \State $base \gets \textsc{ParseBaseSliceFromSentence}(D,q)$
  \If{$\textsc{ApplyConstraints}(D,base,A)\neq\emptyset$} \State \textbf{continue} \EndIf
  \State $dialogue \gets [\ ]$
  \ForAll{$a \in A$}
    \State $slot \gets a.\texttt{column}$
    \State $dialogue.\textsc{append}(\texttt{"ASSISTANT: "} + Q(slot))$
    \State $ans \gets \text{RP}.\textsc{GenerateJSON}(\textsc{RoleplayerPrompt}(q,u,slot))$
    \State $dialogue.\textsc{append}(\texttt{"USER: "} + ans)$
  \EndFor
  \State $x \gets \textsc{AgentPrompt}(q,\ dialogue,\ A,\ \texttt{"All constraints yield no results. Relax EXACTLY ONE."})$
  \State $y^+ \gets \textsc{JSON}(\texttt{relaxed\_constraint}=a^\star)$
  \State $C \gets A \setminus \{a^\star\}$
  \If{$C=\emptyset$} \State \textbf{continue} \EndIf
  \State $ranked \gets [\ ]$
  \ForAll{$c \in C$}
    \State $A_{\setminus c} \gets A \setminus \{c\}$
    \State $R \gets \textsc{ApplyConstraints}(D,base,A_{\setminus c})$
    \State $h \gets \begin{cases}0,&|R|=0\\1,&|R|>0\end{cases}$
    \State $ranked.\textsc{append}((h,|R|,c))$
  \EndFor
  \State $ranked.\textsc{sort}(\text{by }(h\uparrow,\ |R|\uparrow))$
  \State $Negs \gets \textsc{PickTopKWithMix}(ranked,\ K=\textit{max\_negatives\_per\_item})$
  \ForAll{$c^- \in Negs$}
    \State $y^- \gets \textsc{JSON}(\texttt{relaxed\_constraint}=c^-)$
    \State $D_{\text{DPO}} \gets D_{\text{DPO}} \cup \{(x,y^+,y^-)\}$
  \EndFor
\EndFor
\State $\pi_\theta \gets \textsc{LoadModel}(M_0,\ \text{adapter}=\textsc{NewLoRA}())$
\State $\pi_{\text{ref}} \gets \textsc{LoadModel}(M_0)$
\State $\theta_{\text{AG}} \gets \textsc{DPO}(\pi_\theta,\ \pi_{\text{ref}},\ D_{\text{DPO}},\ \beta)$
\State \textsc{SaveAdapter}$(\theta_{\text{AG}})$
\State \Return $(\theta_{\text{RP}},\ \theta_{\text{AG}})$
\end{algorithmic}
\end{breakablealgorithm}

\makeatletter
\@ifundefined{breakablealgorithm}{%
  \newenvironment{breakablealgorithm}
    {%
      \begin{center}
        \refstepcounter{algorithm}%
        \hrule height .8pt depth 0pt \kern 2pt%
        \renewcommand{\caption}[2][\relax]{%
          {\raggedright\textbf{Algorithm~\thealgorithm:} ##2\par}%
          \ifx\relax##1\relax
            \addcontentsline{loa}{algorithm}{\protect\numberline{\thealgorithm}##2}%
          \else
            \addcontentsline{loa}{algorithm}{\protect\numberline{\thealgorithm}##1}%
          \fi
          \kern 2pt\hrule\kern 2pt%
        }%
    }{%
        \kern 2pt\hrule\relax%
      \end{center}
    }
}{}
\makeatother
\makeatother

\begin{breakablealgorithm}
\caption{S-2/3 (any-repair)}
\label{alg:mus2-any-repair}
\footnotesize
\begin{algorithmic}[1]
\Require Target count $\textit{target}$
\Ensure Generated instances $\textit{results}$

\Statex \textbf{Helper:} $\textsc{SoftSatisfaction}(x,p,U)$ returns a soft score in $[0,1]$.

\Procedure{\textsc{GenerateInstances}}{$\textit{target}$}
  \State $\textit{results} \gets \emptyset$
  \State $\textit{used} \gets \emptyset$
  \While{$|\textit{results}| < \textit{target}$}
    \State $U \gets \textsc{SelectUnusedBaseSlice}(\textit{used})$ \Comment{$|U|$ within bounds}
    \If{$U=\texttt{null}$} \State \textbf{break} \EndIf
    \State $P \gets \textsc{BuildCandidatePredicates}(U)$
    \State $(a,b) \gets \textsc{FindDropAnyOnePair}(P)$
      \Comment{$(a\wedge b)=\emptyset,\ a\neq\emptyset,\ b\neq\emptyset,$ prefer distinct attributes}
    \If{$(a,b)=\texttt{null}$} \State \textbf{continue} \EndIf
    \State $\textit{used} \gets \textit{used} \cup \{\textsc{SignatureOf}(U)\}$

    \State $(w_a,w_b) \gets \textsc{SampleNormalizedWeights}()$
    \State $\textit{relaxed} \gets \arg\min\{(a,w_a),(b,w_b)\}$ \Comment{by weight}
    \State $\textit{kept} \gets \{a,b\} \setminus \{\textit{relaxed.predicate}\}$

    \State $C \gets \{\,x \in U:\ x \text{ satisfies all predicates in } \textit{kept}\,\}$
    \If{$C=\emptyset$}
      \State $\textit{results} \gets \textit{results} \cup \{\textsc{Instance}(U,\{a,b\},\{w_a,w_b\},\textit{relaxed},\texttt{null})\}$
      \State \textbf{continue}
    \EndIf

    \ForAll{$x \in C$}
      \State $\textit{score}[x] \gets 0$
      \ForAll{$(p,w) \in \{(a,w_a),(b,w_b)\}$}
        \State $\textit{score}[x] \gets \textit{score}[x] + w \cdot \textsc{SoftSatisfaction}(x,p,U)$
      \EndFor
    \EndFor

    \State $\textit{best} \gets \arg\max_{x\in C}\ \textit{score}[x]$ \Comment{tie-breaker: lowest\_cost}
    \State $\textit{results} \gets \textit{results} \cup \{\textsc{Instance}(U,\{a,b\},\{w_a,w_b\},\textit{relaxed},\textit{best})\}$
  \EndWhile
  \State \Return $\textit{results}$
\EndProcedure

\Procedure{\textsc{BuildCandidatePredicates}}{$U$}
  \State $P \gets \emptyset$
  \ForAll{\textbf{chosen categorical attribute}}
    \ForAll{\textbf{value with sufficient support in $U$}}
      \State $p \gets \textsc{EqualityPredicate}(\textit{value})$
      \If{$p$ covers all of $U$ \textbf{or} none of $U$} \State \textbf{continue} \EndIf
      \State $P \gets P \cup \{p\}$
    \EndFor
  \EndFor
  \ForAll{\textbf{chosen numeric attribute with direction} ($\ge$ or $\le$)}
    \State $T \gets \textsc{QuantileThresholds}(U)$ \Comment{with caps/filters}
    \ForAll{\textbf{threshold} in $T$}
      \State $p \gets \textsc{InequalityPredicate}(\textit{threshold})$
      \If{$p$ covers all of $U$ \textbf{or} none of $U$} \State \textbf{continue} \EndIf
      \State $P \gets P \cup \{p\}$
    \EndFor
  \EndFor
  \State $P \gets \textsc{Deduplicate}(P)$
  \State $P \gets \textsc{FilterBySupport}(P)$
  \State $P \gets \textsc{CapPerAttributeAndTotal}(P)$
  \State \Return $P$
\EndProcedure

\Procedure{\textsc{FindDropAnyOnePair}}{$P$}
  \ForAll{\textbf{many random pairs} $(a,b)$ drawn from $P$}
    \If{\textbf{attributes distinct} \textbf{and} $(a \wedge b)=\emptyset$ \textbf{and} $a\neq\emptyset$ \textbf{and} $b\neq\emptyset$}
      \State \Return $(a,b)$
    \EndIf
  \EndFor
  \ForAll{\textbf{all pairs} $(a,b)$ in $P \times P$}
    \If{\textbf{attributes distinct} \textbf{and} $(a \wedge b)=\emptyset$ \textbf{and} $a\neq\emptyset$ \textbf{and} $b\neq\emptyset$}
      \State \Return $(a,b)$
    \EndIf
  \EndFor
  \State \Return \texttt{null}
\EndProcedure

\Function{\textsc{SoftSatisfaction}}{$x,p,U$}
  \If{$p$ is equality}
    \State \Return $\mathbf{1}[x \text{ matches } p]$
  \ElsIf{$p$ is ``at least''}
    \State \Return $\textsc{NormalizedValue}(x.\textit{attribute}\ \text{over}\ U)$
  \Else
    \State \Return $1 - \textsc{NormalizedValue}(x.\textit{attribute}\ \text{over}\ U)$ \Comment{$p$ is ``at most''}
  \EndIf
\EndFunction
\end{algorithmic}
\end{breakablealgorithm}

\makeatletter
\@ifundefined{breakablealgorithm}{%
  \newenvironment{breakablealgorithm}
    {%
      \begin{center}
        \refstepcounter{algorithm}%
        \hrule height .8pt depth 0pt \kern 2pt%
        \renewcommand{\caption}[2][\relax]{%
          {\raggedright\textbf{Algorithm~\thealgorithm:} ##2\par}%
          \ifx\relax##1\relax
            \addcontentsline{loa}{algorithm}{\protect\numberline{\thealgorithm}##2}%
          \else
            \addcontentsline{loa}{algorithm}{\protect\numberline{\thealgorithm}##1}%
          \fi
          \kern 2pt\hrule\kern 2pt%
        }%
    }{%
        \kern 2pt\hrule\relax%
      \end{center}
    }
}{}
\makeatother

\begin{breakablealgorithm}
\caption{S1 (one unique repair)}
\label{alg:mus2-one-fix}
\footnotesize
\begin{algorithmic}[1]
\Require Target count $\textit{target}$, predicate pool builder \textsc{BuildCandidatePredicates}$(\cdot)$
\Ensure Generated instances $\textit{results}$

\Statex \textbf{Helper:} $\textsc{SoftSatisfaction}(x,p,U)$ returns a soft score in $[0,1]$.

\Procedure{\textsc{GenerateOneFixInstances}}{$\textit{target}$}
  \State $\textit{results} \gets \emptyset$
  \State $\textit{used} \gets \emptyset$
  \While{$|\textit{results}| < \textit{target}$}
    \State $U \gets \textsc{SelectUnusedBaseSlice}(\textit{used})$ \Comment{$|U|$ within bounds}
    \If{$U=\texttt{null}$} \State \textbf{break} \EndIf

    \State $P \gets \textsc{BuildCandidatePredicates}(U)$
    \State $(Q,r) \gets \textsc{FindOneFixQuad}(P)$ \Comment{$Q$ has 4 predicates; $r$ is unique repair index}
    \If{$(Q,r)=\texttt{null}$} \State \textbf{continue} \EndIf
    \State $\textit{used} \gets \textit{used} \cup \{\textsc{SignatureOf}(U)\}$

    \State $W \gets \textsc{AssignWeightsWithUniqueLowest}(Q,r)$
      \Comment{$W[r]$ strictly smallest; normalized}

    \State $\textit{keep} \gets Q \setminus \{Q[r]\}$
    \State $C \gets \{\,x \in U:\ x \text{ satisfies all predicates in } \textit{keep}\,\}$
    \If{$C=\emptyset$} \State \textbf{continue} \Comment{should not occur if constructed correctly} \EndIf

    \ForAll{$x \in C$}
      \State $\textit{score}[x] \gets 0$
      \ForAll{$(p,w) \in \textsc{Zip}(Q,W)$}
        \State $\textit{score}[x] \gets \textit{score}[x] + w \cdot \textsc{SoftSatisfaction}(x,p,U)$
      \EndFor
    \EndFor

    \State $\textit{best} \gets \arg\max_{x \in C}\ \textit{score}[x]$ \Comment{tie-breaker: lowest\_cost}
    \State $\textit{results} \gets \textit{results} \cup \{\textsc{Instance}(U,Q,W,\textit{repair}=Q[r],\textit{recommendation}=\textit{best})\}$
  \EndWhile
  \State \Return $\textit{results}$
\EndProcedure

\Procedure{\textsc{FindOneFixQuad}}{$P$}
  \State $T \gets \emptyset$
  \ForAll{\textbf{triples} $(i,j,k)$ drawn from $P$}
    \If{\textsc{RequireDistinctAttributes} \textbf{and} \textbf{not} all-distinct$(i,j,k)$} \State \textbf{continue} \EndIf
    \State $\textit{bits} \gets \textsc{IntersectBits}(i,j,k)$
    \If{$\textit{bits} \neq 0$}
      \State $T \gets T \cup \{(i,j,k,\textit{bits})\}$ \Comment{triple feasible}
    \EndIf
  \EndFor
  \If{$T=\emptyset$} \State \Return \texttt{null} \EndIf
  \State \textsc{Shuffle}$(T)$

  \ForAll{$(i,j,k,\textit{bits}_{ijk}) \in T$}
    \ForAll{$l$ in indices$(P)$ excluding $\{i,j,k\}$}
      \If{\textsc{RequireDistinctAttributes} \textbf{and} \textbf{not} all-distinct$(i,j,k,l)$} \State \textbf{continue} \EndIf
      \If{$(\textit{bits}_{ijk} \cap \textsc{Bits}(l)) \neq 0$}
        \State \textbf{continue} \Comment{full 4-way must be empty}
      \EndIf
      \If{$\textsc{IntersectBits}(i,j,l) \neq 0$} \State \textbf{continue} \EndIf
      \If{$\textsc{IntersectBits}(i,k,l) \neq 0$} \State \textbf{continue} \EndIf
      \If{$\textsc{IntersectBits}(j,k,l) \neq 0$} \State \textbf{continue} \EndIf

      \State $Q \gets [\textsc{Predicate}(i),\ \textsc{Predicate}(j),\ \textsc{Predicate}(k),\ \textsc{Predicate}(l)]$
      \State $r \gets$ index of $l$ within $Q$ \Comment{dropping $Q[r]$ yields feasible triple}
      \State \Return $(Q,r)$
    \EndFor
  \EndFor
  \State \Return \texttt{null}
\EndProcedure

\Procedure{\textsc{AssignWeightsWithUniqueLowest}}{$Q,r$}
  \State $W \gets$ array of 4 weights
  \State $W[r] \gets \textsc{SampleSmall}()$
  \ForAll{index $t \neq r$}
    \State $W[t] \gets \textsc{SampleLarge}()$
  \EndFor
  \State \textsc{Normalize}$(W)$ \Comment{sum to 1}
  \State \Return $W$
\EndProcedure

\Function{\textsc{SoftSatisfaction}}{$x,p,U$}
  \If{$p$ is equality}
    \State \Return $\mathbf{1}[x \text{ matches } p]$
  \ElsIf{$p$ is ``at least''}
    \State \Return $\textsc{NormalizedValue}(x.\textit{attribute}\ \text{over}\ U)$
  \Else
    \State \Return $1 - \textsc{NormalizedValue}(x.\textit{attribute}\ \text{over}\ U)$ \Comment{$p$ is ``at most''}
  \EndIf
\EndFunction
\end{algorithmic}
\end{breakablealgorithm}

\clearpage

\newtcolorbox{dialoguebox}{
  colback=blue!5,
  colframe=blue!35,
  boxrule=0.6pt,
  arc=2mm,
  left=2mm,right=2mm,top=1.5mm,bottom=1.5mm
}

\newlist{dialoguelist}{itemize}{1}
\setlist[dialoguelist]{leftmargin=*, itemsep=0.35em, label={}, labelsep=0pt}

\subsection{Example Dialogue}
\label{exampledialog}

\begin{dialoguebox}
\begin{dialoguelist}
  \item \textbf{\faUser\ } I am looking for a Chevrolet Extended Cab Pickup with a AUTOMATIC transmission and rear wheel drive.
  \item \textbf{\faRobot\ } Do you have a specific engine fuel type preference for the Chevrolet Extended Cab Pickup?
  \item \textbf{\faUser\ } I’d like it to run on flex-fuel (unleaded/E85), but I’m flexible there if everything else fits. That’s a nice-to-have.
  \item \textbf{\faRobot\ } Do you have a specific MSRP range in mind for the Chevrolet Extended Cab Pickup?
  \item \textbf{\faUser\ } Must-have. I’m pretty firm on keeping MSRP at no more than \$24{,}515, because budget discipline matters most to me.
  \item \textbf{\faRobot\ } Do you have a specific vehicle size preference for the Chevrolet Extended Cab Pickup?
  \item \textbf{\faUser\ } I strongly prefer it to be a Large truck, though I could compromise a little on size if the overall deal is right.
  \item \textbf{\faRobot\ } What is your desired highway MPG for the Chevrolet Extended Cab Pickup, and how important is this to you?
  \item \textbf{\faUser\ } I’d like it to have at least 20 highway MPG, but that’s more of a “nice floor” than a hard line.
\end{dialoguelist}
\end{dialoguebox}

\newcommand{\code}[1]{%
  {\renewcommand{\UrlFont}{\ttfamily\scriptsize}\url{#1}}%
}

\newtcolorbox{evaltablebox}{
  enhanced,
  width=\linewidth,
  colback=blue!5,
  colframe=blue!35,
  boxrule=0.6pt,
  arc=2mm,
  left=1mm,right=1mm,top=1mm,bottom=1mm
}

\subsection{Evaluation}
\label{Eval}
\begin{table}[!htbp]
\centering
\caption{Evaluation metrics used in our experiments}
\label{evaltab}

\begin{evaltablebox}
\footnotesize
\setlength{\tabcolsep}{4pt}
\renewcommand{\arraystretch}{1.25}
\rowcolors{2}{blue!3}{white}
\setlength{\emergencystretch}{2em} 
\sloppy

\begin{tabularx}{\linewidth}{@{}
>{\RaggedRight\arraybackslash}p{3.1cm}
>{\RaggedRight\arraybackslash}p{3.4cm}
>{\RaggedRight\arraybackslash}X
>{\RaggedRight\arraybackslash}X
@{}}
\toprule
\rowcolor{blue!15}
\textbf{Category} & \textbf{Metric} & \textbf{What it measures} & \textbf{How computed} \\
\midrule

Dialogue completion & Avg \#Slots Provided &
How many constraint slots the agent was asked to fill &
\code{mean(len(slots_provided_to_agent))} \\

Dialogue completion & Avg \#Constraints Parsed &
How many constraints the system actually extracted &
\code{mean(len(parsed_additional_constraints))} \\

Dialogue completion & Slot Completion Rate &
Fraction of provided slots that resulted in a parsed constraint &
\code{total_parsed/total_slots} \\

Dialogue completion & Per-slot Completion Rate &
Which slots are frequently missed &
\code{for_each_slot:parsed_count/asked_count} \\

Planning outcome & SAT (No Relaxation) Rate &
Full-query feasibility success &
\code{count(status==SAT_no_relaxation)/N} \\

Planning outcome & SAT (After Relaxation) Rate &
Success after relaxing $\ge 1$ constraint &
\code{count(status==SAT_after_relaxation)/N} \\

Planning outcome & UNSAT Rate &
Failed even after trying relaxation &
\code{count(status==UNSAT_even_after_relaxation)/N} \\

Planning outcome & Recommendation Produced Rate &
Whether the system produced any car &
\code{count(recommended_car!=null)/N} \\

Oracle agreement & Relax Match Rate &
Whether relaxed constraint matches oracle relaxed constraint &
\code{match(relaxed_constraint,oracle_relaxed)_over_comparable_instances} \\

Oracle agreement & Top-1 Car Match \textbar\ Relax-Correct &
Car match, only when relaxation matches oracle &
\code{match(car_key(agent),car_key(oracle))_on_relax_correct_subset} \\

\bottomrule
\end{tabularx}
\end{evaltablebox}
\end{table}

\newtcolorbox{promptbox}{
  enhanced,
  colback=blue!5,
  colframe=blue!35,
  boxrule=0.6pt,
  arc=2mm,
  left=2mm,right=2mm,top=1.5mm,bottom=1.5mm
}

\lstdefinestyle{prompt}{
  basicstyle=\ttfamily\footnotesize,
  columns=fullflexible,
  breaklines=true,
  breakatwhitespace=true,
  keepspaces=true,
  showstringspaces=false,
  aboveskip=0.5em,
  belowskip=0.5em
}

\subsection{Infeasibility Task Prompts}
\label{taskprompts}

\begin{promptbox}
\textbf{ROLEPLAYER\_SYS (System).}
You are \texttt{ROLEPLAYER\_USER}. You only know the \emph{persona} and the \emph{base query}.  
Answer using \textbf{only} information stated or clearly implied in the persona:
(i) if the persona mentions any relevant value for the asked slot, \textbf{state it explicitly};
(ii) say ``no preference'' \textbf{only} if the persona contains no information for that slot;
(iii) if the question is vague, still provide the most relevant constraint value(s);
(iv) do not invent details.

\medskip
\textbf{AGENT\_ASK\_SYS (System).}
You are \texttt{CLARIFYING\_AGENT}. Ask \textbf{one} short question to learn the user's requirement for the given slot.  
No relaxations; no multi-part questions.

\medskip
\textbf{AGENT\_EXTRACT\_SYS (System).}
You are a strict JSON generator. From the user's answer, extract:
(1) the slot constraint (or \texttt{null}), and (2) an importance weight in $[0,1]$.  
Output \textbf{only JSON}.
\end{promptbox}

\begin{promptbox}
\textbf{agent\_question (User prompt template)}
\begin{lstlisting}[style=prompt]
Base query: {base_query}

Conversation so far:
{hist_txt}

Slot to clarify: {slot}

Ask ONE concise question to determine the user's requirement for {slot}.
\end{lstlisting}
\end{promptbox}

\begin{promptbox}
\textbf{roleplayer\_answer (User prompt template)}
\begin{lstlisting}[style=prompt]
Base query the user started with:
{base_query}

Persona (all you know):
{persona}

Assistant asked:
{question}

Answer as the user, using only persona information.
\end{lstlisting}
\end{promptbox}

\begin{promptbox}
\textbf{extract\_constraint\_and\_weight (User prompt template)}
\begin{lstlisting}[style=prompt]
Slot: {slot}
User answer: {user_answer}

Return ONLY JSON:
{
  "constraint": null OR {"column":"{slot}","op":"=="|">="|"<=","value":string|number},
  "weight": number
}

Rules:
- No preference => constraint:null and weight 0.0.
- Numeric slots (Year, city mpg, highway MPG, MSRP):
  >= for minimums, <= for budgets/maximums, == only if explicitly exact.
  value must be an integer.
- Categorical slots (Engine Fuel Type, Vehicle Size, Model):
  op == and value is a string.
- weight in [0,1]:
\end{lstlisting}
\end{promptbox}

\newtcolorbox{personabox}{
  enhanced,
  colback=blue!5,
  colframe=blue!35,
  boxrule=0.6pt,
  arc=2mm,
  left=2mm,right=2mm,top=1.5mm,bottom=1.5mm
}

\subsection{Example Personas (MUS-4 unique repair, MUS-4 any repair, MUS-2 any repair)}
\label{sec:example_personas}

\begin{personabox}
\textbf{S1 (unique repair).} I’m looking for a Lexus IS 250 sedan with an automatic transmission and rear-wheel drive, because I like that sporty-but-composed feel. The model is my top priority—I really want an IS 250 and I’m not very willing to compromise on that. I also strongly prefer it to be Midsize, since that size feels like the perfect balance for me. I’m aiming for at least 21 city MPG, and I care about that, but I can be a bit flexible if the car is otherwise a great match. For fuel, I’d like regular unleaded, but I’m extremely flexible there and can easily compromise on it.
\end{personabox}

\begin{personabox}
\textbf{S3 (any repair).} I want a Mercedes-Benz sedan with an automatic transmission and rear-wheel drive, because I’m after that classic smooth luxury feel. The biggest must-haves for me are that it takes premium unleaded (required) and that it’s specifically an E-Class, since those are core to what I’m looking for and I’m not really willing to compromise on them. I’d strongly prefer it to be 2016 or newer, but I could bend a little if everything else is perfect. I’m aiming for at least 23 city MPG, though I’m flexible on that and can compromise if the driving experience and model match are right.
\end{personabox}

\begin{personabox}
\textbf{S2 (any repair).} I’m looking for a Buick sedan with an automatic transmission and front-wheel drive, and I’m really set on it being a Verano. The model matters most to me, so I’m not very willing to compromise on Verano specifically. I’d like it to run on premium unleaded (required), but I’m more flexible there and can compromise on fuel type if the Verano match is right.
\end{personabox}

\lstdefinestyle{prompt}{
  basicstyle=\ttfamily\footnotesize,
  columns=fullflexible,
  breaklines=true,
  breakatwhitespace=true,
  keepspaces=true,
  showstringspaces=false,
  aboveskip=0.4em,
  belowskip=0.4em
}

\subsection{Persona Generation Prompts and Example Personas}
\label{persona}

\begin{promptbox}
\textbf{Persona generator prompt (concise).}
\begin{lstlisting}[style=prompt]
You are a persona generator.

Input: JSON (one object or an array). Each object includes:
- base_query_sentence
- constraint_weights (constraints + weights; may include is_unique_repair)
- chosen_relaxation (optional)

Task: For each object, write ONE first-person persona paragraph (3 to 5 sentences) that:
1) restates what the user wants (from base_query_sentence),
2) mentions every constraint in constraint_weights exactly once,
3) conveys priority + flexibility:
   - highest: must-have, not willing to compromise
   - high: strongly prefer / care a lot
   - medium: aim for it, can compromise a bit
   - low: nice-to-have, very flexible
   If chosen_relaxation exists, describe it as the MOST flexible and place it last.

Style: friendly, natural, with a mild motivation in sentence 1
(e.g., sedan=comfortable daily car; SUV=practical; pickup=workhorse; coupe=fun).
No numeric weights. Do not mention "weight", "unique repair", or "relaxation policy".

Output: JSON ONLY.
- If input is an array: output a JSON array of strings (same order).
- If input is a single object: output one JSON string.

INPUT:
{{INPUT_JSON_HERE}}
\end{lstlisting}
\end{promptbox}

\end{document}